\documentclass[10pt,twocolumn,letterpaper]{article}

\usepackage[pagenumbers]{cvpr} 

\definecolor{cvprblue}{rgb}{0.21,0.49,0.74}
\usepackage[pagebackref,breaklinks,colorlinks,allcolors=cvprblue]{hyperref}

\usepackage{booktabs} 
\usepackage{enumitem}
\usepackage{diagbox}
\usepackage{amsmath}
\usepackage{amssymb}
\usepackage{xspace}
\usepackage{soul}
\usepackage{media9}
\usepackage{multirow}
\usepackage{graphicx}  
\usepackage{pifont} 
\usepackage{bbding}
\usepackage[accsupp]{axessibility}
\usepackage{wasysym}
\newcommand{\cmark}{\ding{51}}%
\newcommand{\xmark}{\ding{55}}%
\newcommand{\sysname}{EffectMaker\xspace}

\graphicspath{figs/}

\def\confName{CVPR}
\def\confYear{2026}

\title{\sysname: Unifying Reasoning and Generation for \\ Customized Visual Effect Creation}

\author{
Shiyuan Yang$^{1,2,\dagger}$ \quad
Ruihuang Li$^{1,\dagger}$ \quad
Jiale Tao$^{1}$ \quad
Shuai Shao$^{1,*}$ \quad
Qinglin Lu$^{1,\ddagger}$ \quad
Jing Liao$^{2,\ddagger}$ \\
\\
$^{1}$Tencent Hunyuan \quad
$^{2}$City University of Hong Kong \\
\\
\small
$^{\dagger}$Equal contribution.
$^{*}$ Project lead.
$^{\ddagger}$Corresponding authors.
}

\begin{document}

\twocolumn[{%
    \maketitle
    \vspace{-3em} 
    \begin{center}
        \centering
        \includegraphics[width=\textwidth]{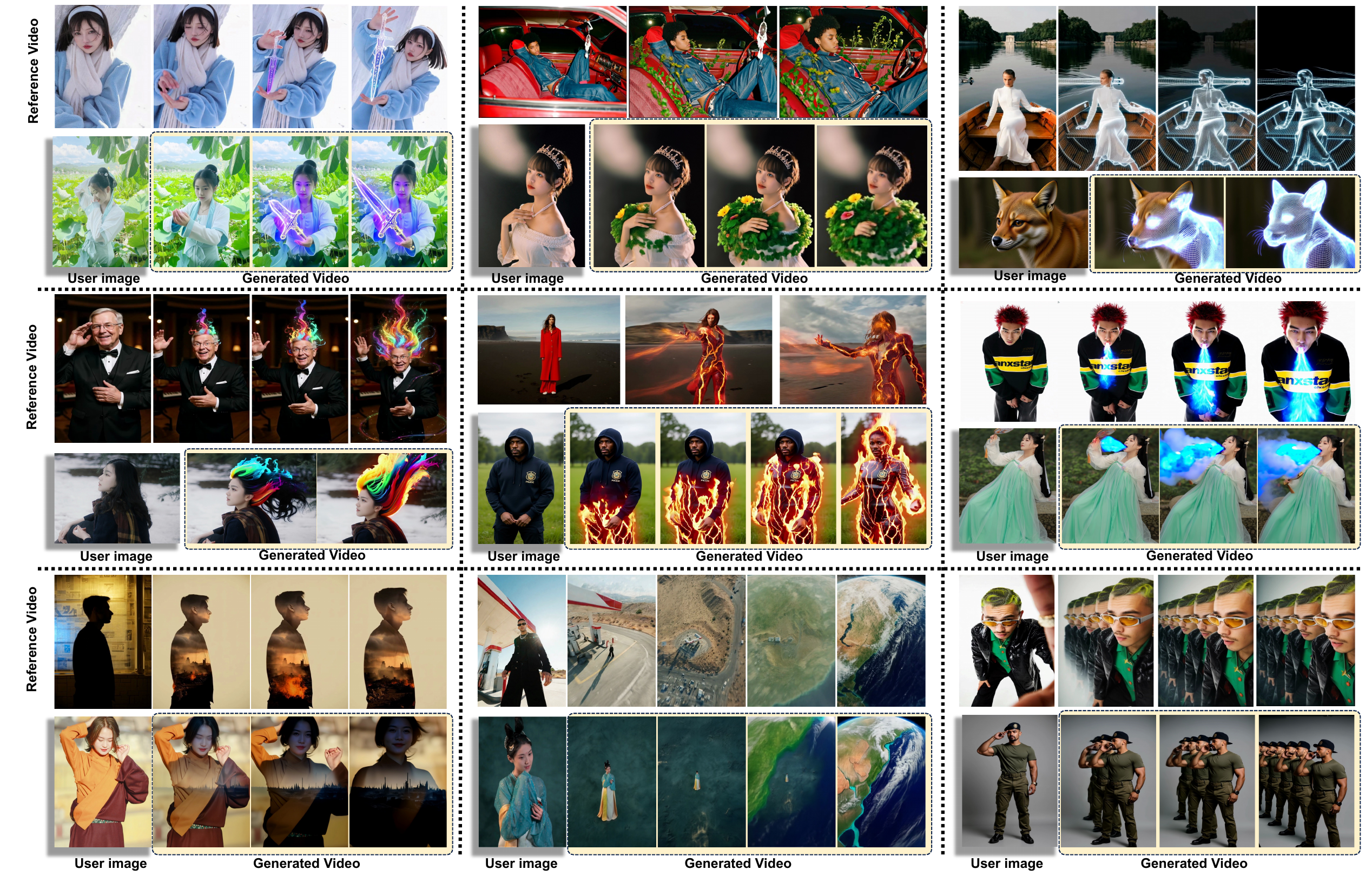} 
        \captionof{figure}{Given a reference video with visual effect (top row in each grid), and a user-specified target image (wrapped by shadow box), our \sysname transfers the reference effect to user image to create vivid video (bottom row in each grid) with the same effect pattern.}
        \label{fig:teaser}
    \end{center}
    \vspace{0.2em} 
}]

\begin{abstract}
Visual effects (VFX) are essential for enhancing the expressiveness and creativity of video content, yet producing high-quality effects typically requires expert knowledge and costly production pipelines. Existing AIGC systems face significant challenges in VFX generation due to the scarcity of effect-specific data and the inherent difficulty of modeling supernatural or stylized effects. Moreover, these approaches often require per-effect fine-tuning, which severely limits their scalability and generalization to novel VFX.
In this work, we present \sysname, a unified reasoning–generation framework that enables reference-based VFX customization. \sysname employs a multimodal large language model to interpret high-level effect semantics and reason about how they should adapt to a target subject, while a diffusion transformer leverages in-context learning to capture fine-grained visual cues from reference videos. These two components form a semantic–visual dual-path guidance mechanism that enables accurate, controllable, and effect-consistent synthesis without per-effect fine-tuning.
Furthermore, we construct EffectData, the largest and high-quality synthetic dataset containing 130k videos across 3k VFX categories, to improve generalization and scalability. Experiments show that \sysname achieves superior visual quality and effect consistency over state-of-the-art baselines, offering a scalable and flexible paradigm for customized VFX generation. Project page: \url{https://effectmaker.github.io}.
\end{abstract}
    
\section{Introduction}
\label{sec:intro}

Visual effects (VFX) are designed to enhance the visual expressiveness and aesthetic appeal of a video by introducing artistic or physical augmentations beyond the real world. They play a crucial role in film, advertising, and gaming, serving as a creative medium to enrich narrative experience and visual immersion. However, producing high-quality VFX typically requires professional expertise and expensive production pipelines, making it one of the most resource-intensive components in modern video creation. As generative AI technologies evolve, leveraging AIGC (AI-generated content) for VFX generation holds great promise for reducing production costs while enabling accessible, creative, and controllable video effect design.

Despite remarkable advances in video generative models, current AIGC systems remain underexplored for visual effect generation. Although large-scale diffusion-based video models trained on massive real-world datasets can generate visually realistic content, they struggle with supernatural, non-physical, and exaggerated effects — which are inherently out-of-domain from real-world video distributions. Moreover, VFX-specific data are scarce and costly to obtain, limiting systematic research in this direction. Existing attempts often rely on small, closed-set VFX categories and require either fine-tuning a dedicated LoRA for each effect \cite{vfxcreator} — which is inefficient and hard to scale, or training a mixture of LoRAs on a limited effect set \cite{omnieffect}, which lacks generalization to open-set unseen effects.

A natural alternative for open-set VFX generation is to describe target effects via text prompts, as in text-to-video systems. However, visual effects are often abstract, multi-layered, and stylistically complex, making them difficult to capture precisely with language alone. Even for professional designers, textual descriptions may not always be a good choice as text often fail to convey the nuanced texture, motion dynamics, and atmosphere of a desired effect. In practice, designers often rely on visual references—transferring the “look and feel” of one effect into another visual context—an intuitive and visually grounded process that textual prompting cannot achieve. Therefore, beyond the data scarcity challenge, a key technical problem lies in accurately extracting and transferring effects from reference videos.

In this work, we present \sysname, a unified reasoning–generation framework designed for effect-driven video synthesis.
On the reasoning side, a multimodal large language model (MLLM) first comprehends the effect's high-level semantics from the reference, and then reasons about how to adaptively apply this effect to the novel subject in user's image. This frees users from explicitly describing complex visual effects. On the generation side, a diffusion transformer (DiT)~\cite{dit} leverages in-context learning to capture fine-grained visual details from the reference. Together, these two components form a semantic–visual dual-path guidance mechanism that jointly steers the model toward semantically aligned and visually consistent VFX synthesis. By framing effect creation as a reference-based task and leveraging an MLLM for enhanced visual understanding and reasoning, our method offers an intuitive, scalable, and highly controllable alternative to previous LoRA-based approaches, eliminating the need for effect-specific fine-tuning.

To address the VFX data scarcity, we construct EffectData, the largest VFX dataset to date, comprising over 130k videos across 3k diverse effect categories, including atmospheric, transformation, stylistic effects. EffectData expands the effect category by an order of magnitude compared with existing datasets and offers paired annotations. We will release the dataset to support future research in VFX generation and editing.
We validate \sysname against state-of-the-art effect generation methods. Extensive experiments demonstrate that our approach achieves superior visual quality and effect consistency.

In summary, our main contributions are as follows:

\begin{itemize}
  \item We propose \sysname, a reference-based framework for VFX customization. It unifies multimodal understanding and controllable generation in a single model to apply an effect from a reference video onto a target image.
  \item We introduce semantic–visual dual-path guidance mechanism, which combines MLLM-based understanding with a Diffusion Transformer's in-context learning ability, enabling accurate and controllable VFX transfer.
 \item We construct EffectData, the largest high-quality VFX dataset to date, containing 3k effect classes, providing a valuable resource for future research on VFX generation.
\end{itemize}

\section{Related work} \label{sec:rw}

\paragraph{General video generation.}
Recent advances in diffusion models have significantly advanced text-to-video (T2V) generation. 
Early T2V approaches extended 2D image diffusion models into the temporal domain using U-Net–based backbones equipped with spatio-temporal attention or motion modules~\cite{animatediff, videoldm, svd}, enabling controllable short-video generation.
More recently, the field has shifted toward Diffusion Transformers (DiTs)~\cite{dit}, which offer greater scalability and stronger generative capacity. Prominent DiT-based systems such as CogVideo~\cite{cogvideox}, Wan~\cite{wan}, HunyuanVideo~\cite{hunyuanvideo}, Sora~\cite{sora}, and Veo~\cite{veo3} have achieved unprecedented realism and fidelity in open-domain video generation, which has greatly benefited down-stream applications in creative visual creation~\cite{dav,zsy_flexiact,xiao2026binary,li2025syncnoise,wu2025insvie,chen2025fast}.
%

\paragraph{Generation with understanding.}

With the rapid development of MLLMs in the field of multi-modal understanding~\cite{zsy_narrative, zsy_logo,lin2025jarvisart, lin2025jarvisevo}, increasing attention has been devoted to unifying vision understanding and visual generation within a single reasoning–generation framework. Existing approaches can be grouped into three paradigms.
The first extends auto-regressive LLMs to visual domains by predicting discrete visual tokens~\cite{chameleon, emu3, januspro, unitoken, zhang2025duetsvg}, but typically suffers from limited fidelity and weak spatial coherence.
The second, LLM–diffusion hybrids~\cite{metaquery, seedx, omnigen2, chen2025postercraft}, uses the LLM for high-level reasoning while delegating image synthesis to an external diffusion decoder, achieving a practical balance between semantic capability and visual quality.
The third paradigm explores unified architectures~\cite{transfusion, showo, bagel} that merge LLM and diffusion into a single transformer, offering tighter semantic–visual coupling but with high computational cost and limited scalability for video.
Following recent understanding–generation frameworks~\cite{veggie, instructx, mindomni}, we adopt the second paradigm for its lightweight yet effective integration of understanding and generation in reference-based visual effect synthesis.

\paragraph{Visual effect generation.}
Visual effect generation remains a relatively underexplored problem in the video generation community. The pioneering AIGC-based method MagicVFX~\cite{magicvfx} synthesizes effects by directly copying pixel-level content from a reference video to a target video within a user-specified region, followed by noise injection for refinement. However, this copy–paste manner lacks flexibility when the reference and target scenes differ significantly, and it requires extensive manual adjustment.
VFXCreator~\cite{vfxcreator} enables effect transfer by fine-tuning a separate LoRA for each effect type, achieving reasonable results for single effects but suffering from poor scalability and generalization to unseen ones. Omni-Effects~\cite{omnieffect} improves scalability via a mixture-of-LoRA (LoRA-MoE) strategy trained on 55 effect categories, yet the limited effect diversity constrains domain generalization.
Two concurrent works, Video-as-Prompt~\cite{vap} and VFXMaster~\cite{vfxmaster}, adopt a reference-based paradigm similar to ours and demonstrate better generalization to unseen effects. Nevertheless, their models lacks reasoning ability, relies on carefully and manually-crafted effect prompts, making them unfriendly for interactive use.
Moreover, existing VFX datasets cover limited range of effect categories (dozens to a few hundred), whereas our EffectData increases this scale by an order of magnitude. It also provides per-video annotations including labels, captions, and instructions. A statistic comparison of VFX-related dataset is shown in \cref{tab:data}.

\begin{table}[t]
\centering
\caption{Comparison of existing VFX datasets. Our dataset significantly expands the number of effect classes and provides detailed captions and editing instructions.}
\setlength{\tabcolsep}{2pt} 
\renewcommand{\arraystretch}{0.9} 
\resizebox{0.45\textwidth}{!}{
\begin{tabular}{l|ccccc}
\toprule
\textbf{Dataset} & \textbf{\# Classes} & \textbf{\# Videos} &  \textbf{Label}&\textbf{Caption} & \textbf{Instruction} \\
\midrule
VFX-307\cite{magicvfx}     & --   & 190 &  \xmark &\xmark & \xmark \\
Open-VFX\cite{vfxcreator}     & 52   & 634 &  \cmark &\xmark & \xmark \\
Omni-VFX\cite{omnieffect}     & 55   & 1700 &  \cmark &\xmark & \xmark \\
HiggisField\cite{higgsfield}  & 145  & 1060 &  \cmark &\cmark & \xmark \\
VAP-Data\cite{vap}            & 100  & 90k&  \cmark &\xmark & \xmark \\
\textbf{EffectData (Ours)} & \textbf{3000} & \textbf{130k} &  \cmark &\cmark & \cmark \\
\bottomrule
\end{tabular}}

\label{tab:data}
\end{table}

\section{Method} \label{sec:method}

\subsection{Overview}
In this work, we propose \sysname, a unified reference-based visual effect generation model that seamlessly integrates the sophisticated visual effect understanding and reasoning capabilities of MLLMs with the in-context generative power of video DiT model. We also introduce EffectData, the largest paired VFX dataset with comprehensive annotations. Our task is formulated as follows: given a single reference VFX video and a target image, \sysname analyzes and comprehends the VFX in the reference video and transfers them to the target image, generating visually compelling videos with the same stylistic effects.
%



\begin{figure*}[t]
  \centering
   \includegraphics[width=\linewidth]{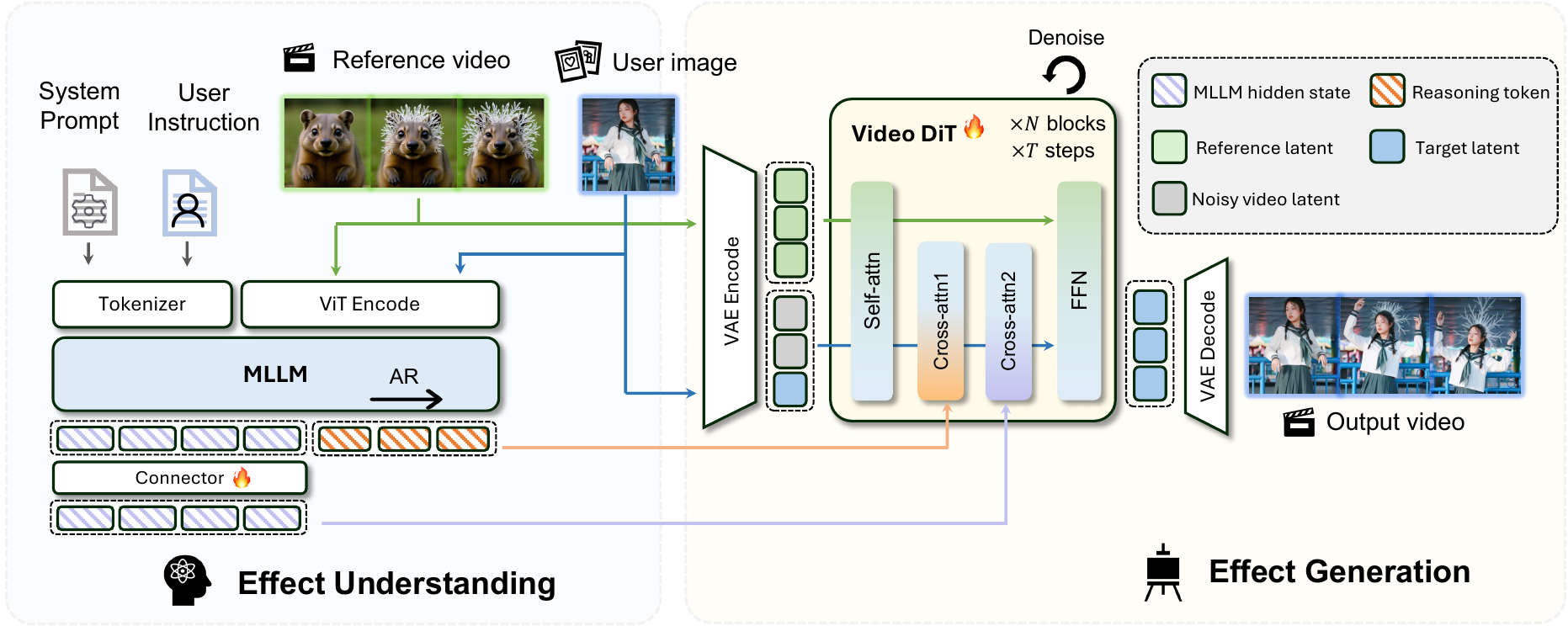}
    \caption{Overview of our model architecture. Given a reference VFX video and a target image, on the reasoning side, an MLLM extracts high-level semantic cues of the reference video, providing abstract effect descriptions that serve as semantic guidance. On the generation side, a video DiT model leverages in-context generation to capture fine-grained visual details from the reference, and generates a target video with consistent visual effect.}
    \label{fig:pipeline}
\end{figure*}

An overview of our model architecture is illustrated in \cref{fig:pipeline}. Our model consists of two main components: on the understanding side, an MLLM model is responsible for semantically understanding and reasoning about the reference visual effects; On the generation side, a video DiT synthesizes the target video conditioned on the extracted effect representations.

\subsection{Effect understanding}
On the understanding side, our MLLM is initialized from the Qwen3-VL-8B model \cite{qwen3vl}. On the input side, it encodes multimodal information including the system prompt, user instruction, reference video, and user image. The user instruction guides the MLLM to first analyze the visual effects presented in the reference video, then examine the content of the user image, and reason how the effect should adapt to a new target, especially those with major differences in shape, and finally imagine and describe the appearance of the target image after transferring the specified visual effect.
On the output side, to fully exploit the model’s multimodal understanding and reasoning capabilities, we extract two complementary types of features: \textit{semantic-understanding features} and \textit{semantic-reasoning features}. The semantic-understanding features are obtained from the hidden states of the last layer of the MLLM, encoding rich multimodal representations of the reference inputs. However, this feature only captures the current input semantics through a single forward pass and does not involve any reasoning about how the reference effect should be applied to the user’s image.
To complement this, we further extract semantic-reasoning features from the auto-regressively predicted text token sequence. These features summarize the model’s previous understanding and encapsulate its reasoning process to infer the user-desired final outcome. The two feature types are then jointly used as conditioning signals for the video DiT.
Since the feature spaces of the MLLM and the DiT are not aligned, we introduce a light connector module to bridge this modality gap. The overall workflow is illustrated in the left part of \cref{fig:pipeline}.

\begin{figure*}[t]
  \centering
   \includegraphics[width=1.0\linewidth]{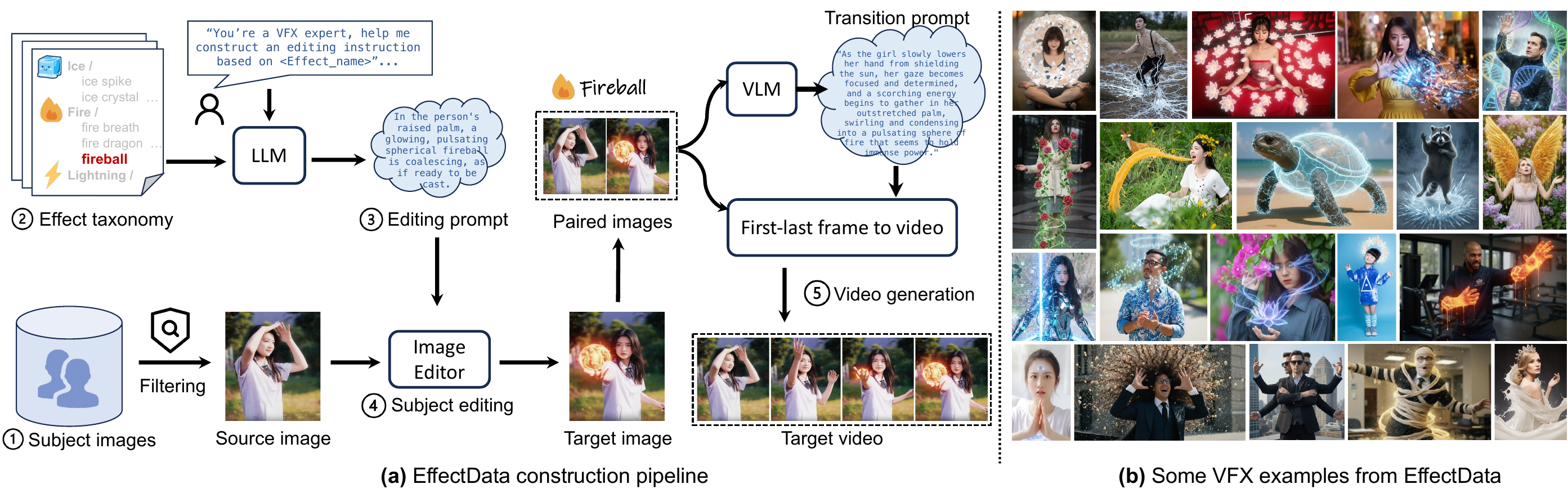}
    \caption{(a) Illustration of our EffectData construction pipeline. (b) Some examples from the EffectData dataset.}
    \label{fig:data}
\end{figure*}

\subsection{Effect generation} \label{sec:method.gen}
On the generation side, we employ an image-to-video base model to transfer the reference effect onto a user-provided subject image, thereby synthesizing a dynamic visual effect video. Specifically, we adopt Wan2.2-TI2V-5B \cite{wan} as our DiT-based backbone. To achieve faithful effect cloning, we introduce a semantic–visual dual-conditioning strategy that injects reference effect information from both the semantic and visual levels.

\paragraph{Semantic conditioning via decoupled cross-attention.}
The semantic-level conditioning is derived from the understanding stage described above, encompassing both MLLM's understanding and reasoning features. A straightforward fusion strategy is to concatenate these two modalities and feed them into the DiT’s cross-attention layers. However, we found that such a direct combination compromises the model’s representational capacity.
To better preserve modality-specific information, we employ a decoupled cross-attention mechanism. Concretely, reasoning features are encoded using the original T5 text encoder of DiT and injected through the standard cross-attention layers. For understanding features, we introduce an additional cross-attention branch that processes them independently. Both branches share the same query representation but use distinct key and value projections for the textual and visual features. The outputs from the two branches are then fused together by direct feature addition. To prevent unintended semantic interference with the reference stream, the cross-attention is performed exclusively between the target video tokens and the semantic conditions.



\paragraph{Visual conditioning via in-context learning.}
While semantic conditioning provides global guidance on what effect to generate, it lacks the fine-grained spatial and temporal details required for visually faithful effect cloning. To address this limitation, we further leverage the in-context learning capability of DiT for visual-level conditioning, following prior studies~\cite{fulldit, camclonemaster}.
Specifically, both the reference and target videos are encoded into latent representations using a shared VAE encoder. After patchifying and flattening, the reference and target latents are concatenated along the sequence dimension and jointly processed by the DiT blocks.
Notably, we modify the self-attention into a dual-stream scheme, where reference and target tokens are projected through separate query, key, and value projections to disentangle their representation spaces, while still allowing bi-directional attention across the combined token sequence. The attention operation is formulated as:
\begin{equation}
\begin{aligned}
O_r &= \mathrm{SA}\!\left(Q_r, [K_r; K_t], [V_r; V_t]\right) \\
O_t &= \mathrm{SA}\!\left(Q_t, [K_r; K_t], [V_r; V_t]\right)
\end{aligned},
\end{equation}
where $\mathrm{SA}(\cdot)$ denotes the self-attention operation, $Q_r, K_r, V_r$ and $Q_t, K_t, V_t$ are the query, key, and value features of the reference and target video tokens, respectively. $[\cdot;\cdot]$ denotes concatenation along the sequence dimension. $O_r$ and $O_t$ are the output features for the reference and target streams, which are then jointly processed by subsequent feed-forward layers.

\paragraph{Biased RoPE.}
To differentiate between the target and reference videos, we employ a biased 3D Rotary Position Embedding (RoPE) for the reference video. Specifically, we align the spatial components of the RoPE to those of the target video, while introducing a constant offset to the temporal dimension. This temporal bias separates the positional encoding spaces of the two videos, ensuring a safe margin to prevent interference between their RoPE.

\begin{figure*}[t]
  \centering
  \includegraphics[width=1.0\linewidth]{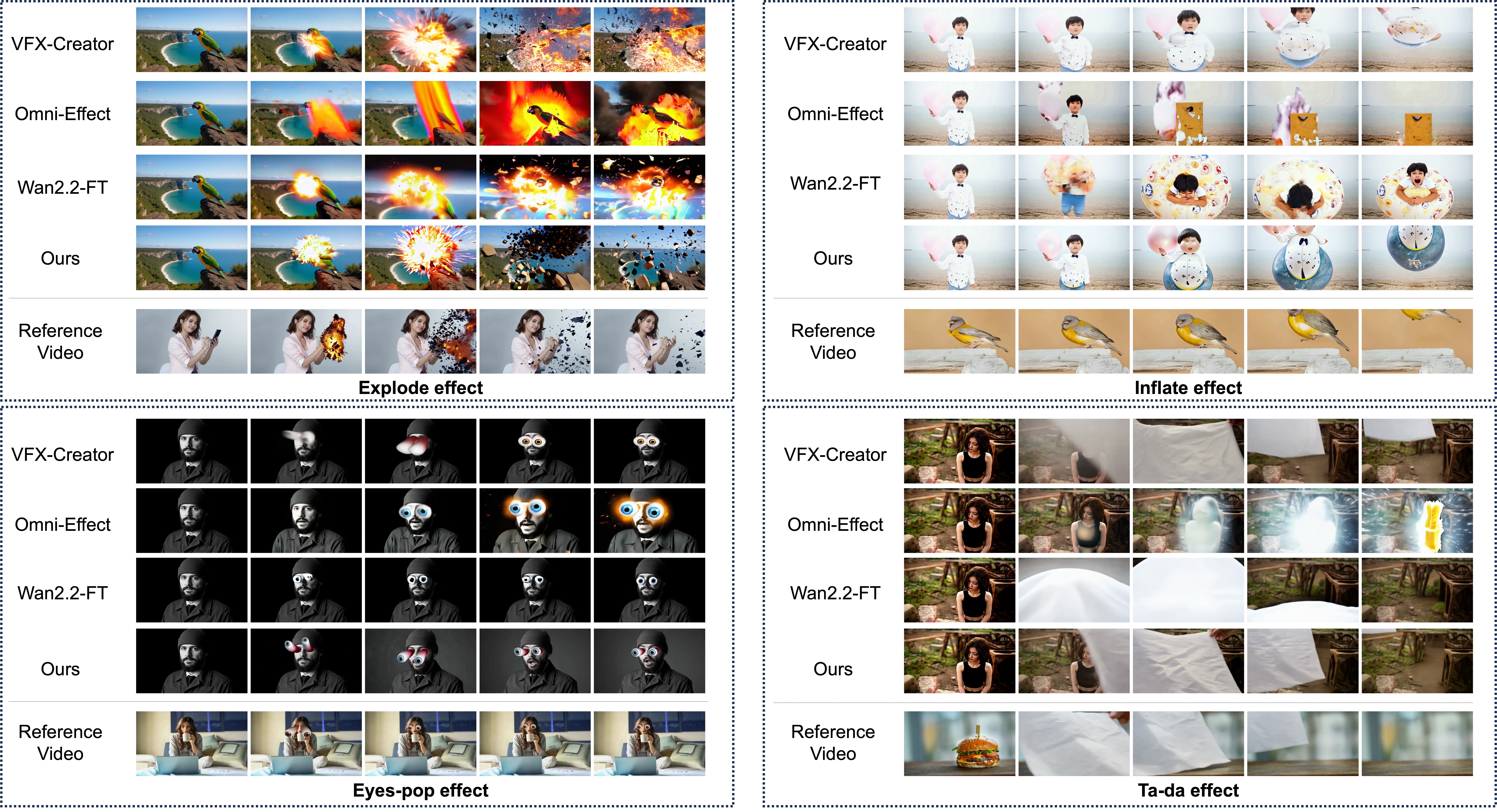}
    \caption{Qualitative comparison with related baselines on OpenVFX dataset.}
    \label{fig:quali}
\end{figure*}

\begin{table*}[t]
\centering
\caption{Quantitative comparison with related baselines on the OpenVFX dataset. VQ=visual quality, MQ=motion quality, TA=text alignment, CAS=class alignment score. All metrics are model-based; higher values indicate better performance.}
\setlength{\tabcolsep}{2pt} 
\renewcommand{\arraystretch}{0.9} 
\resizebox{\textwidth}{!}{
\begin{tabular}{llccccccccccccccc}
\toprule
\textbf{Metrics} & \textbf{Methods} & Cake-ify & Crumble & Decapitate & Deflate & Dissolve & Explode & Eye-pop & Harley & Inflate & Levitate & Melt & Squish & Ta-da & Venom & Avg. \\
\midrule
\multirow{5}{*}{\textbf{VQ}↑} 
& VFX-Creator & 2.71 & 2.41 & 2.68 & 2.33 & 1.69 & 1.38 & 2.13 & 0.97 & 2.41 & 2.50 & 1.96 & 1.77 & 2.28 & 1.62 & 2.06 \\
& Omni-Effect & 3.11 & 2.49 & 2.62 & 2.39 & \textbf{1.93} & 1.84 & 2.23 & 1.48 & 2.55 & 2.27 & 2.20 & 2.27 & 2.59 & 1.82 & 2.27 \\
& Wan2.2-FT & 2.15 & 1.42 & 2.63 & 2.15 & 1.64 & 0.91 & 2.23 & 1.71 & 1.89 & 2.12 & 1.74 & 2.11 & 2.04 & 1.69 & 1.89 \\
& Ours & \textbf{3.39} & \textbf{2.94} & \textbf{3.01} & \textbf{2.99} & \textbf{1.93} & \textbf{2.23} & \textbf{2.44} & \textbf{2.17} & \textbf{3.19} & \textbf{3.70} & \textbf{2.53} & \textbf{2.79} & \textbf{3.99} & \textbf{2.46} & \textbf{2.84} \\
\midrule
\multirow{5}{*}{\textbf{MQ}↑} 
& VFX-Creator & -0.05 & -0.15 & -0.05 & 0.13 & -0.17 & -0.35 & 0.11 & -0.26 & -0.14 & 0.48 & -0.16 & -0.41 & \textbf{0.77} & -0.21 & -0.03 \\
& Omni-Effect & 0.12 & -0.27 & -0.08 & 0.02 & -0.09 & -0.33 & 0.04 & -0.28 & -0.11 & -0.04 & -0.14 & -0.37 & -0.09 & -0.34 & -0.14 \\
& Wan2.2-FT & 0.22 & -0.16 & \textbf{0.79} & 0.41 & \textbf{0.06} & -0.30 & 0.45 & \textbf{0.27} & 0.04 & 0.28 & 0.09 & \textbf{-0.02} & 0.63 & 0.01 & 0.20 \\
& Ours & \textbf{0.59} & \textbf{0.02} & 0.11 & \textbf{0.51} & -0.06 & \textbf{-0.13} & \textbf{0.73} & 0.25 & \textbf{0.09} & \textbf{0.65} & \textbf{0.26} & -0.05 & 0.52 & \textbf{0.04} & \textbf{0.25 }\\
\midrule
\multirow{5}{*}{\textbf{TA}↑} 
& VFX-Creator & -2.63 & 1.54 & -1.82 & -3.38 & -3.50 & 1.44 & 0.90 & \textbf{1.51} & -2.95 & -1.29 & -0.58 & \textbf{0.99} & -3.36 & 0.26 & -0.92 \\
& Omni-Effect & -2.23 & \textbf{1.80} & -1.99 & -3.10 & -3.32 & \textbf{1.77} & 1.27 & \textbf{1.51} & -1.71 & -1.04 & -0.22 & 0.87 & -2.43 & \textbf{0.28} & -0.61 \\
& Wan2.2-FT & \textbf{-0.56} & 1.43 & -2.20 & -2.51 & -1.38 & 0.87 & 0.75 & -1.22 & -0.04 & -1.03 & -1.50 & 0.94 & -2.62 & -0.74 & -0.70 \\
& Ours & -1.89 & 1.40 & \textbf{-1.60} & \textbf{-2.31} & \textbf{-0.55} & 1.11 & \textbf{1.29} & 1.00 & \textbf{0.41} & \textbf{-0.47} & \textbf{-0.17} & 0.80 & \textbf{-2.35} & -0.08 & \textbf{-0.24} \\
\midrule
\multirow{5}{*}{\textbf{CAS}↑} 
& VFX-Creator & \textbf{4.00} & \textbf{5.00} & \textbf{5.00} & \textbf{5.00} & \textbf{5.00} & 4.20 & \textbf{5.00} & \textbf{5.00} & \textbf{5.00} & 1.00 & \textbf{5.00} & \textbf{5.00} & 1.80 & 4.80 & 4.34 \\
& Omni-Effect  & 2.80 & \textbf{5.00} & 4.00 & 2.80 & 4.80 & \textbf{4.80} & \textbf{5.00} & \textbf{5.00} & \textbf{5.00} & 4.00 & 4.60 & \textbf{5.00} & 3.80 & \textbf{5.00} & 4.40 \\
& Wan2.2-FT    & 2.20 & \textbf{5.00} & 0.00 & \textbf{5.00} & \textbf{5.00} & 3.00 & 4.40 & 4.20 & 3.00 & 4.60 & 1.00 & 4.00 & 3.80 & \textbf{5.00} & 3.59 \\
& Ours         & 3.60 & 4.60 & \textbf{5.00} & 4.20 & \textbf{5.00} & 3.80 & \textbf{5.00} & \textbf{5.00} & 4.60 & \textbf{4.80} & \textbf{5.00} & \textbf{5.00} & \textbf{4.20} & \textbf{5.00} & \textbf{4.63} \\
\bottomrule
\end{tabular}}
\label{tab:quanti}
\end{table*}

\subsection{Data construction} \label{sec:method.data}

Existing VFX video datasets are scarce in both scale and diversity. To overcome this, we build a data synthesis pipeline and construct a large-scale, high-quality paired VFX data. As shown in \cref{fig:data}, the detailed steps are as follows:



\noindent\textbf{Step 1: Subject collection.}
We first construct a subject dataset mainly composed of human portraits and animal images, collected from internal sources and public datasets PPR10K~\cite{ppr10k}. We preprocess the data by filtering out images containing text, multiple or unclear subjects.

\noindent\textbf{Step 2: VFX taxonomy.} 
We build systematic VFX taxonomy by defining a series of orthogonal attribute sets including VFX elements (\eg, ice, fire, magic), geometric patterns (\eg, particle, wave, ring), and attachment regions (\eg, face, arms, full body). This structured taxonomy supports compositional generation of diverse effect classes.

\noindent\textbf{Step 3: Instruction generation.} 
For each VFX class, we prompt an LLM to generate multiple editing instructions describing how to transform the source image into its effect-enhanced counterpart.

\noindent\textbf{Step 4: Subject editing.} 
Given a source image from Step~1 and its corresponding instruction from Step~3, we employ image editing model to synthesize the target image with the desired effect. 

\noindent\textbf{Step 5: Video generation.}
For each source-target image pair, we then prompt an MLLM to describe the dynamic transition between the two images, and feed this prompt along with the frame pair into first-last-frame-to-video model to synthesize temporally coherent VFX videos.

\section{Experiment} \label{sec:exp}

\subsection{Experiment setup} \label{sec:exp.setup}

\paragraph{Dataset.} Our datasets are collected from multiple sources, including EffectData synthesized using our proposed data pipeline (\cref{sec:method.data}), supplemented by additional samples from OpenVFX dataset~\cite{vfxcreator} and the Higgsfield website~\cite{higgsfield}. 
%

\paragraph{Implementation details.}
During training, the reference and target videos are randomly sampled from the same VFX class. To reduce computational cost, we temporally downsample each reference video to 17 frames. Each frame is then resized such that the shorter side equals 448 pixels. For the target video, we set the frame length to 81 frames with the shorter side fixed to 704 pixels, and the longer side is proportionally adjusted to maintain the aspect ratio of the user-provided first frame. We train our model for approximately 50k steps on 32 NVIDIA H20 GPUs using the Adam optimizer \cite{adam} with a learning rate of $2\times10^{-5}$.

%


\paragraph{Metrics.}
We evaluate model performance from two complementary perspectives: video quality and effect consistency. For quality assessment, we employ reward-model-based metrics from VideoAlign~\cite{videoalign}, including Visual Quality (VQ) score, which evaluates appearance fidelity, and the Motion Quality (MQ) score, which measures temporal smoothness and motion realism. For effect consistency, we use the Text Alignment (TA) score from VideoAlign, which quantifies the alignment between generated videos and textual descriptions. We further leverage Gemini 2.5 API to rate a 0-5 Effect Class Alignment Score (CAS) and an Effect Reference Alignment Score (RAS) that reflect how well the effect in the generated video matches the intended VFX class name and reference VFX.

\paragraph{Baselines.}
We compare our method with previous state-of-the-art VFX generation methods including VFXCreator \cite{vfxcreator} and Omni-Effects \cite{omnieffect}, we also finetuned our base model Wan2.2-TI2V-5B on our dataset as a baseline method, referred as Wan2.2-FT.

\subsection{Qualitative results} \label{sec:exp.quali}
\paragraph{Close-set comparison.}
Since the baseline methods VFXCreator and Omni-Effects were trained on the OpenVFX dataset, which contains a closed set of visual effects, we also trained our model on the same dataset and selected the common categories as our testing benchmark for fair comparison. In \cref{fig:quali}, we provide side-by-side qualitative results of different methods. Notably, the reference video in the last row is solely provided for our method only. It is evident from the \textit{Inflate} and \textit{Ta-da} cases that our method produces more accurate and visually consistent effects compared to the Wan2.2-FT and Omni-Effect. For instance, in the \textit{Inflate} case, both Omni-Effect and Wan2.2-FT failed to generate the expected inflated and floating boy, whereas our model achieved a realistic and coherent transformation. Although VFXCreator achieves comparable visual quality in most cases, it requires finetuning per-effect LoRA models, making it less flexible and scalable than our feed-forward approach. Benefiting from strong visual understanding and in-context learning, our method also successfully transfers spatial and temporal effect patterns from the reference video, as seen with the explosion flames and flying debris in the \textit{Explode} effect.

\begin{figure}[t]
  \centering
  \includegraphics[width=1.0\linewidth]{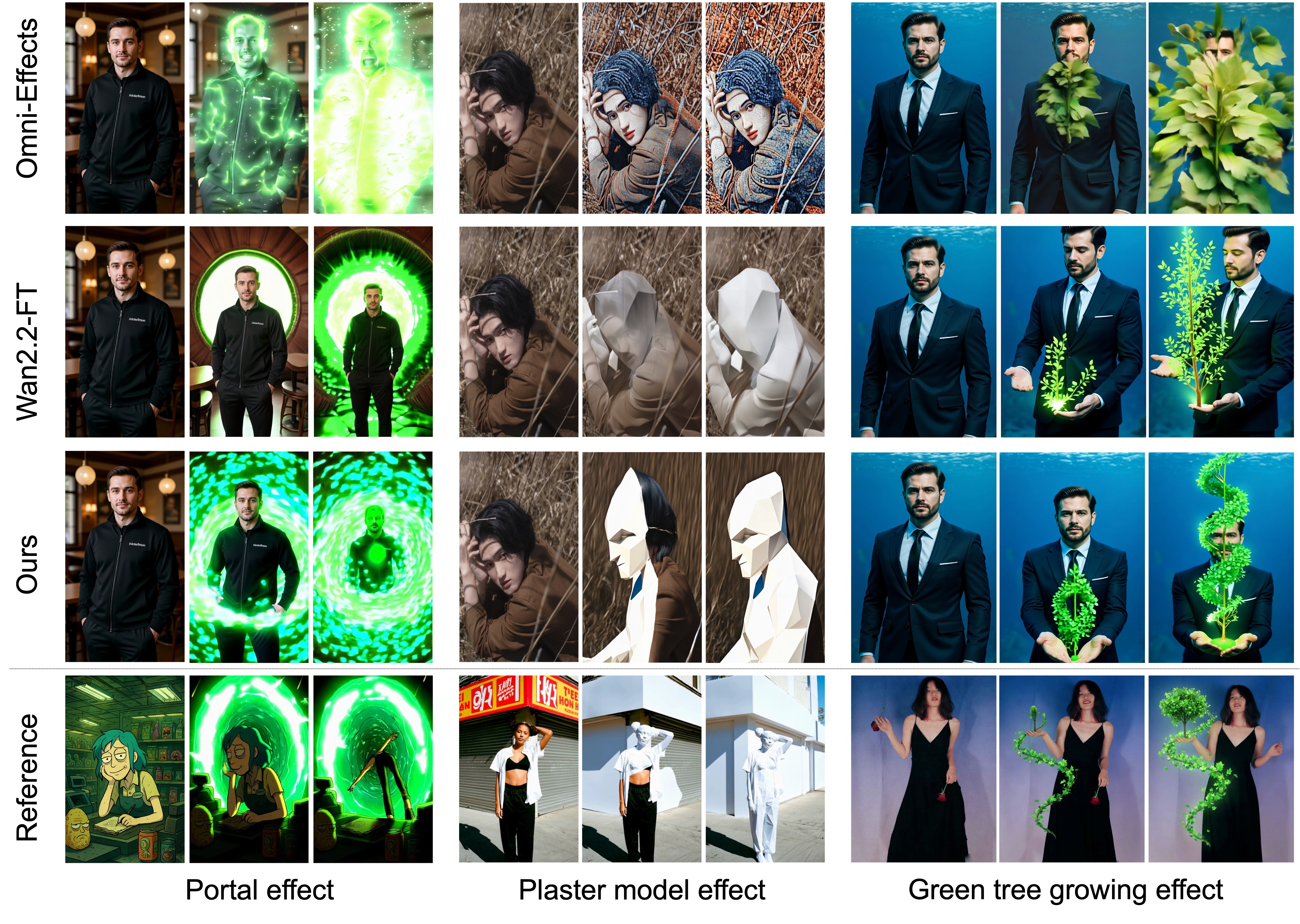}
    \caption{Qualitative comparison on unseen visual effects.}
    \label{fig:open_set}
\end{figure}

\paragraph{Open-set comparison.}
To evaluate the generalization capability of our model, we further compare it with Omni-Effects and Wan2.2-FT on open-set visual effects that were not included in the training set. As illustrated in \cref{fig:open_set}, we present several examples of unseen effects, including \textit{portal}, \textit{plastic model}, and \textit{green tree glowing}. It is evident that Omni-Effects fails to generate plausible effects for unseen categories, while Wan2.2-FT can produce roughly correct patterns but with noticeably lower similarity to the reference effect. This suggests that text-only conditioning struggles to capture the fine-grained visual cues and temporal dynamics of complex effects. In contrast, our method, guided by reference videos, successfully reproduces these unseen effects with higher consistency relative to the reference.

\begin{figure}[t]
  \centering
  \includegraphics[width=1.0\linewidth]{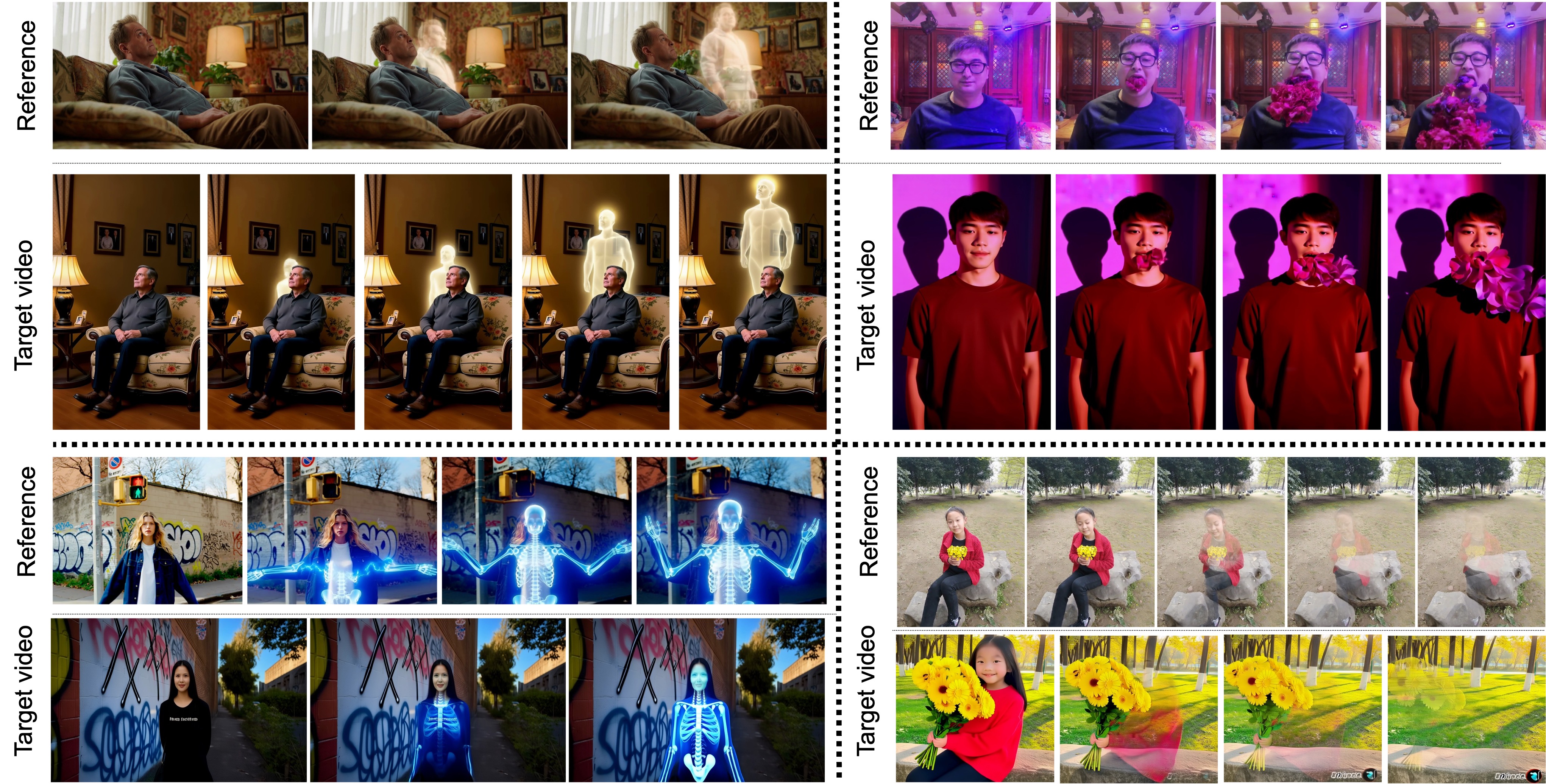}
    \caption{Reference rephrasing results. Given only a reference effect video, our model reproduces the target video with similar visual content and effect.}
    \label{fig:t2v}
\end{figure}

\paragraph{Reference rephrasing.}
Since our framework builds upon the TI2V model, which inherently supports both I2V and T2V generation, it naturally extends to generating effect videos without an input image. In this setting, we leverage the MLLM to describe both the content as well as the VFX in the reference video, and then employ the DiT to render a new video that reproduces the content and effect. We refer to this application as reference rephrasing. We show several examples in \cref{fig:t2v}. The results demonstrate that our semantic–visual dual-path guidance mechanism can effectively rephrase the reference video cues.

\subsection{Quantitative results} \label{sec:exp.quanti}
Following Omni-Effect~\cite{omnieffect}, we conduct a quantitative comparison across 14 visual effect classes from the OpenVFX dataset. For each class, we test on 10 subject images and report the average scores for four metrics: VQ, MQ, TA, and CAS, as summarized in \cref{tab:quanti}. The results demonstrate that our method not only achieves superior visual and motion quality but also surpasses existing state-of-the-art VFX generation methods in terms of effect alignment and responsiveness. Notably, the comparison with the Wan2.2-FT baseline further highlights the advantage of the reference-based paradigm over purely text-driven generation, as the former contributes more effectively to modeling dynamic and complex visual patterns.

\subsection{Ablation study} \label{sec:exp.abl}

\begin{figure}[t]
  \centering
  \includegraphics[width=1.0\linewidth]{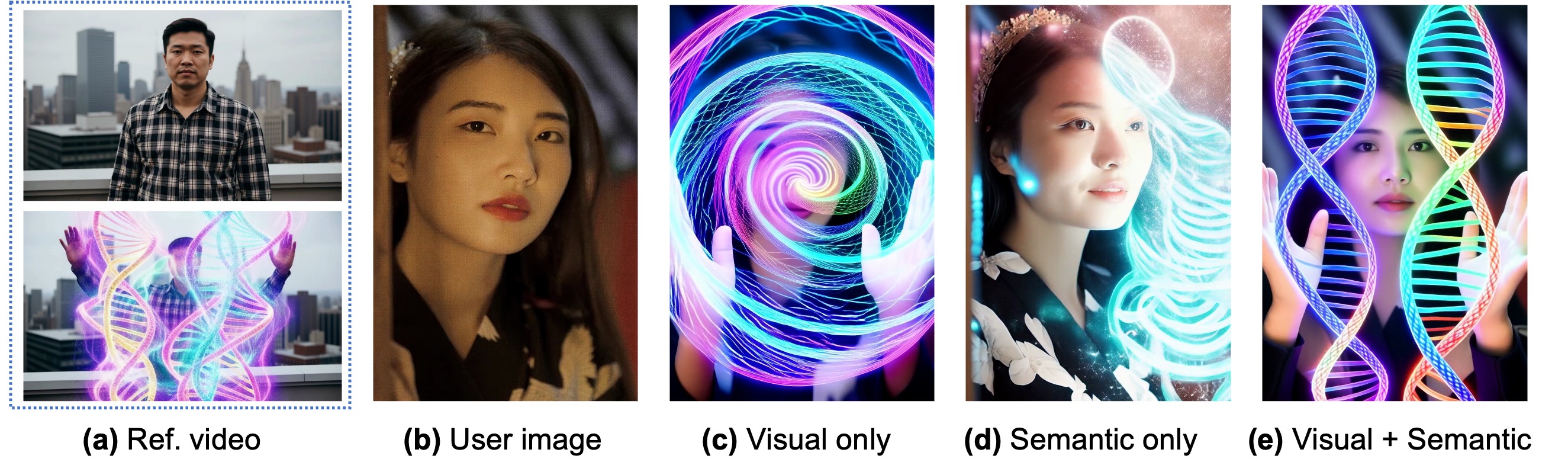}
    \caption{Ablation on different types of conditioning.}
    \label{fig:abl-cond}
\end{figure}

\paragraph{Conditioning design.}
To verify our dual-conditioning strategy, \ie semantic conditioning and visual in-context conditioning, we conduct an ablation study and report the quantitative results in \cref{tab:abl-cond}, along with qualitative comparisons in \cref{fig:abl-cond}. 
As shown, using either semantic conditioning or visual in-context conditioning alone leads to inferior performance. When conditioned solely on the visual reference video, the model tends to replicate only low-level appearance cues, such as color and texture, while failing to capture complex effect structures (\eg, it cannot reproduce the DNA-like geometry as seen in \cref{fig:abl-cond}(c)). In contrast, relying only on semantic conditioning yields semantically correct effects but lacks fine-grained spatial details.
Notably, combining both conditioning achieves the most faithful  performance, demonstrating the effectiveness of our proposed semantic-visual conditioning strategy.

\begin{table}[htbp]
\centering
\small
\caption{Ablation study on different conditioning types.}
\resizebox{0.85\linewidth}{!}{
\begin{tabular}{l|ccccc}
\toprule
Conditioning & \textbf{VQ}$\uparrow$ & \textbf{MQ}$\uparrow$  &\textbf{TA}$\uparrow$& \textbf{CAS}$\uparrow$ & \textbf{RAS}$\uparrow$ \\
\midrule
Semantic& 2.78& 0.16&1.06& 4.20& 3.84\\
Visual& 2.48& 0.12&-0.38& 2.24& 1.48\\
Semantic+Visual& \textbf{2.92}& \textbf{0.21}&\textbf{1.24}& \textbf{4.40}& \textbf{4.16}\\
\bottomrule
\end{tabular}}
\label{tab:abl-cond}
\end{table}

\paragraph{Data scaling.}
We further examine the impact of scaling up the training data on the model’s generalization capability. Specifically, we train our model with two different orders of magnitude of effect classes (\ie 100, and 1000) and evaluate them on open-set effects. The results are reported in \cref{tab:abl2}. As shown, increasing the data scale consistently improves both quality and consistency metrics. We attribute this improvement to the broader domain coverage provided by a larger training set, which enables better interpolation and extrapolation when the model encounters new effects.

\begin{table}[t]
\small
\centering
\caption{Effect of scaling up training set on model performance.}
\label{tab:abl2}
\resizebox{0.85\linewidth}{!}{
\begin{tabular}{l|ccccc}
\toprule
\# VFX Class & \textbf{VQ} $\uparrow$ & \textbf{MQ} $\uparrow$ & \textbf{TA} $\uparrow$ & \textbf{CAS} $\uparrow$ & \textbf{RAS} $\uparrow$ \\
\midrule
100  & 2.76 & 0.13 &  0.94 & 3.76 & 3.22 \\
1000 & \textbf{2.89} & \textbf{0.19} &  \textbf{1.21} & \textbf{4.22}& \textbf{4.04}\\
\bottomrule
\end{tabular}}
\end{table}



\paragraph{Attention design.} We compare our dual-stream attention design mentioned in \cref{sec:method.gen} with a single-stream variant where reference and target tokens share the same projection matrices. We find that switching to single-stream attention degrades the effect transfer capability, with TA decreasing from 1.24 to 0.81, CAS from 4.40 to 3.30, and RAS from 4.16 to 2.84. This is likely because reference and target tokens lie in heterogeneous (clean vs.\ noisy) domains, where separate projections better handle the distribution gap.

\subsection{User study} \label{sec:exp.userstudy}
\begin{figure}[t]
  \centering
  \includegraphics[width=1.0\linewidth]{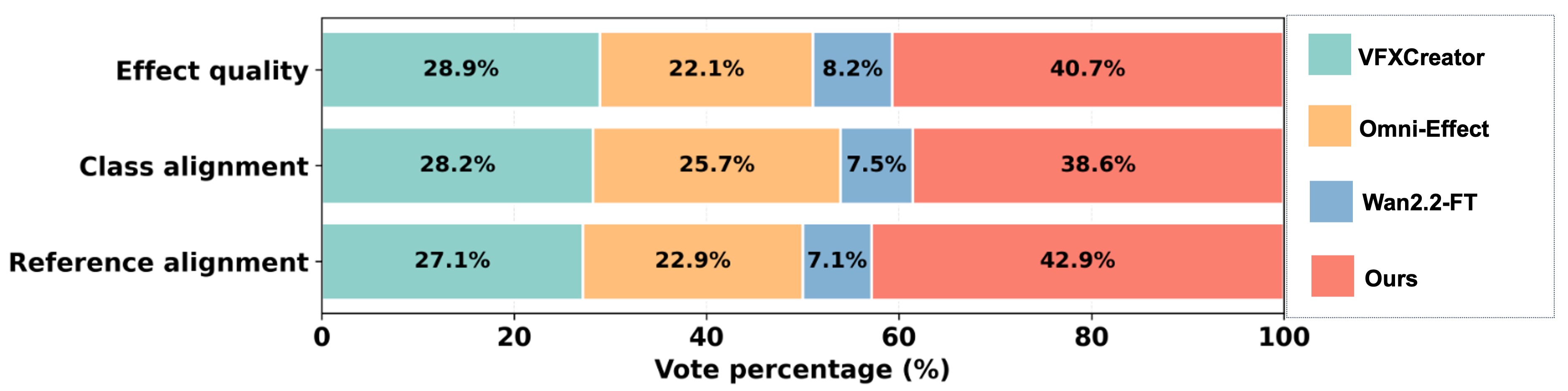}
    \caption{Results of human preference percentage.}
    \label{fig:user_study}
\end{figure}

We conduct a user study on OpenVFX results, involving 30 participants and 28 questions (two per effect). Participants compared side-by-side video results with the effect name and reference video, and selected their preferred result based on three criteria: effect quality, class alignment, and reference alignment. The display order was randomized to reduce bias. As shown in \cref{fig:user_study}, our method achieves the highest preference rates across all criteria, demonstrating superior perceptual quality and effect consistency.

\section{Conclusion} \label{sec:conclusion}
We present \sysname, a reasoning–generation framework for customized visual effect cloning. By integrating the advanced reasoning capability of MLLM with the in-context learning ability of DiT, \sysname achieves superior performance in replicating complex reference effects across diverse scenarios. In addition, we introduce EffectData, the largest and high-quality paired VFX dataset to date, which provides systematic captions and serves as a valuable resource for future research and applications in art, gaming, and advertising communities.
\paragraph{Limitation.} Despite the promising results, \sysname still has several limitations. First, our approach may struggle with extremely complex effects involving rapid or large motions, primarily due to the base model's capacity bound. Second, the reliance on synthetic training data may introduce biases and may not fully capture the diversity and realism of real-world VFX scenarios (\eg, noisy, ambiguous, or multiple overlapping effects). Future work will explore stronger base models and incorporate diverse real-world data sources to further improve generalization and realism.

\newpage

{
\small
\bibliographystyle{ieeenat_fullname}
\bibliography{main}
}

\clearpage
\appendix
\renewcommand{\thesection}{\Alph{section}}

\section{Additional implementation details} \label{sec:sup.impl}

\paragraph{Multi-resolution training.}
Since our framework does not require the reference video and the target video to share the same spatial resolution, we adopt a multi-resolution training strategy to improve robustness to variable input sizes. For the target video, we standardize the shorter side to 704 pixels, while the longer side is adaptively scaled to the nearest multiple of 32—consistent with the VAE’s downsampling factor (but no more than 1280 pixels), ensuring minimal distortion of the original aspect ratio. To reduce computational overhead, we downsample the reference video such that its shorter side is resized to 448 pixels, with the longer side similarly adjusted to a multiple of 32 while capped at 832 pixels.

\paragraph{Timestep sampling.}
For timestep sampling, instead of uniform timestep sampling, we adopt a semi-logit-normal distribution similar to the strategy used in Stable Diffusion 3~\cite{sd3}. Specifically, the timestep $t$ is sampled as follows:
\begin{equation}
    t \sim \left\lfloor T \cdot \frac{1}{1 + e^{-1.5 z}} \right\rfloor, 
    \qquad z \sim \mathcal{N}(0, 1)
    \label{eq:logit_timestep}
\end{equation}
where $T$ denotes the total number of timesteps. This scheme biases the sampling density toward the middle timesteps, which has been shown to outperform uniform sampling in rectified flow-based diffusion models.

\paragraph{Timestep embedding.}
During the forward noising process at timestep $t$, we keep all reference tokens and the first-frame tokens of the target video noise-free. For these clean tokens, we always apply the zero timestep embedding. Noise is added only to the non–first-frame tokens of the target video, and we use the corresponding timestep embeddings for these noised tokens.

\paragraph{DiT sampling.}
The DiT model takes as input the concatenated latent sequence consisting of the target image latent (encoded from the user-provided image), the noise latent (sampled from Gaussian noise), and the encoded reference latent. Conditioned on the semantic features extracted by the MLLM, the DiT progressively denoises this sequence and predicts the target video using flow-matching–based sampling \cite{fm}. We adopt the UniPC sampler~\cite{unipc} with 30 denoising steps and apply classifier-free guidance with a scale of 6~\cite{cfg}. The noise schedule shift parameter is set to 16.

\paragraph{Computational cost and efficiency}
Under the aforementioned configuration, the MLLM-based understanding stage requires 2 seconds, while the DiT-based generation stage takes 2 minutes on an H20 GPU. The end-to-end inference consumes 48 GB of GPU memory. The memory can be further reduced if applying optimization techniques (\eg model sharding, CPU offloading, sequence parallelism, quantization \etc).

\begin{figure}[h]
  \centering
   \includegraphics[width=1.0\linewidth]{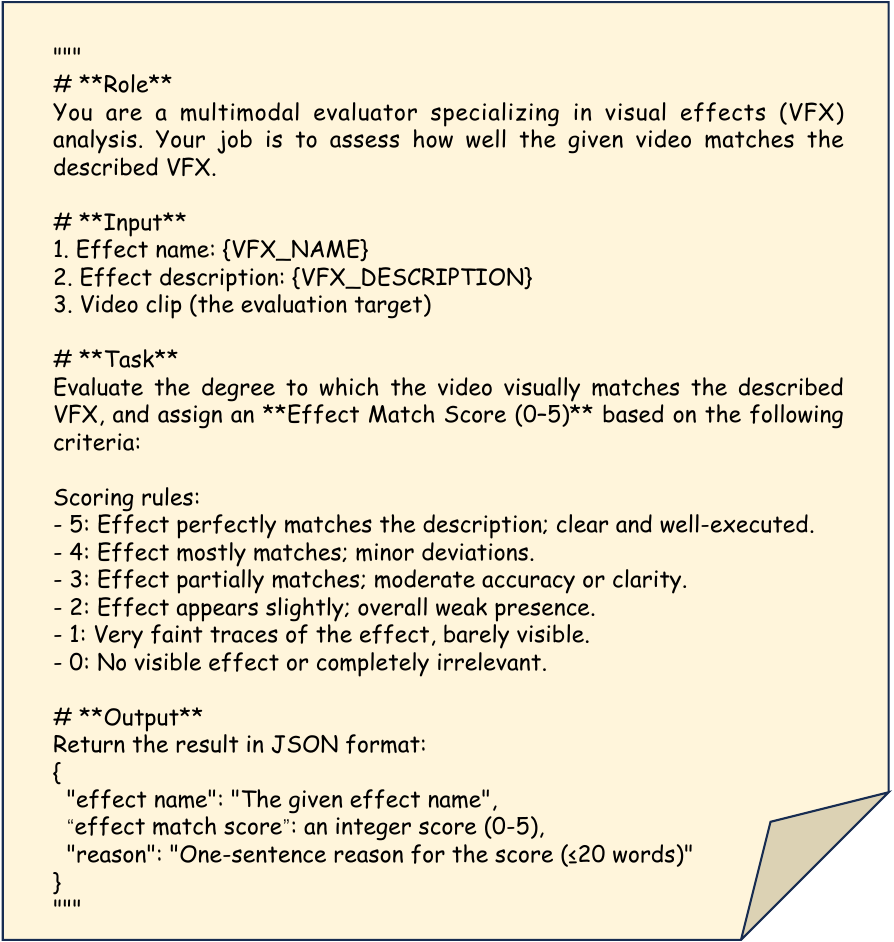}
    \caption{Prompt template for automatic Effect Class Alignment (CAS) Scoring.}
    \label{fig:sup.cas_score}
\end{figure}

\begin{figure*}[h]
  \centering
   \includegraphics[width=0.75\linewidth]{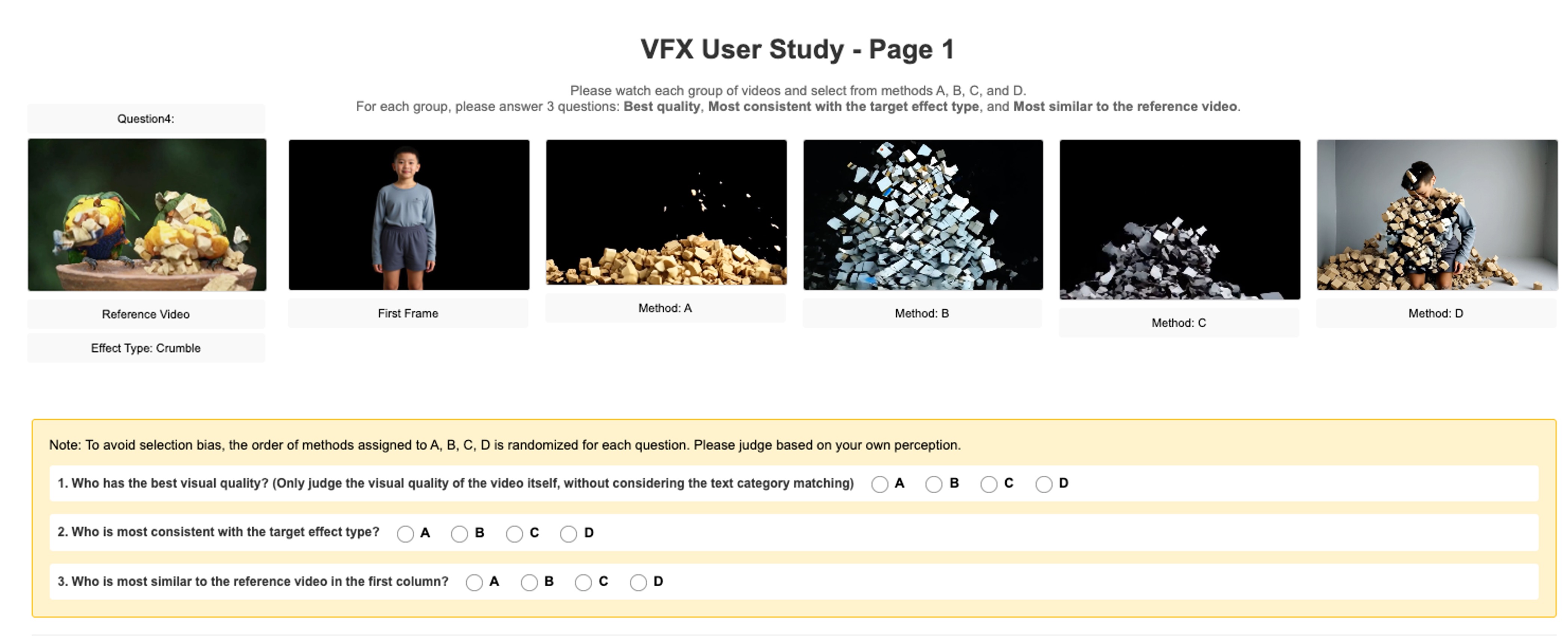}
    \caption{Screen shot of user study questionnaire.}
    \label{fig:questionnare}
\end{figure*}

\begin{figure*}[h]
  \centering
   \includegraphics[width=0.75\linewidth]{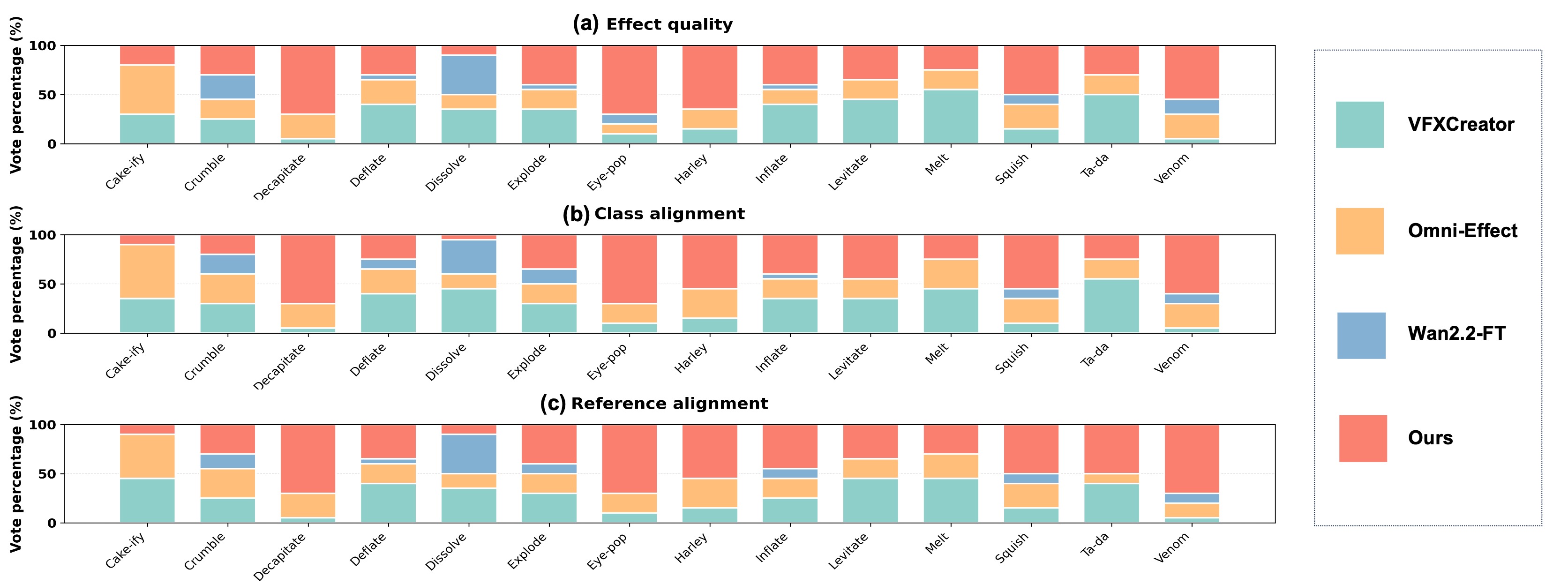}
    \caption{Detailed human preference percentages for each VFX class and each evaluation dimension.}
    \label{fig:user_study_detail}
\end{figure*}

\paragraph{Metrics.}
To evaluate video generation quality, we employ VQ (Visual Quality), MQ (Motion Quality), and TA (Text Alignment) as primary metrics. Derived from the most recent VideoAlign model \cite{videoalign}, a multi-dimensional reward model built on the Qwen2-VL backbone that was trained on 182k human-annotated triplets using a Bradley-Terry-with-Ties (BTT) objective, allowing it to robustly learn relative preferences and "tie" conditions from professional annotators. These metrics are designed to systematically mirror human judgment across distinct dimensions. VQ assesses the static visual fidelity of the video frames, focusing on clarity, aesthetic creativity, and logical reasonableness, independent of the textual context. MQ exclusively evaluates dynamic characteristics, measuring motion smoothness, physical plausibility, and temporal stability. Finally, TA serves as a context-aware metric that gauges the semantic consistency between the generated video and the input prompt, verifying the accurate representation of subjects, actions, and environments. Together, these indicators provide a comprehensive assessment that aligns closely with human perception

In addition to the above metrics, we further introduce the Effect Class Alignment Score (CAS) and Reference Alignment Score (RAS) to assess how well a generated video conforms to the intended visual effect. Specifically, we feed the generated video together with the target effect name (or reference frames) into Gemini2.5 \cite{gemini}, instructing the model to evaluate the degree of alignment between the video and the specified effect class and to assign a score from 0 to 5. Here we show the  CAS scoring prompt in \cref{fig:sup.cas_score}.

\paragraph{User study details}
We conducted a user study to evaluate the perceptual quality of different baselines across the 14 effect categories in the OpenVFX dataset. The study included 28 questions (two per effect) and involved 30 participants. For each question, participants were shown side-by-side video results generated by different methods, along with the effect class name and a reference video. They were asked to choose their preferred result according to three criteria: (1) Effect quality: the overall visual fidelity judged purely by subjective perception; (2) Class alignment: how well the generated effect matches the given class name; and (3) Reference alignment: how closely the result resembles the reference video. To reduce selection bias, the display order in each question was randomly shuffled. We show a screenshot of our user study questionnaire in \cref{fig:questionnare}.

We show detailed human preference percentages for each VFX class and each evaluation dimension in \cref{fig:user_study_detail}. The results show that for most of effect clasess, our method consistently achieved the highest or comparable preference rates across all evaluation dimensions, demonstrating its superior perceptual quality and effect consistency.



\begin{figure}[t]
  \centering
   \includegraphics[width=1.0\linewidth]{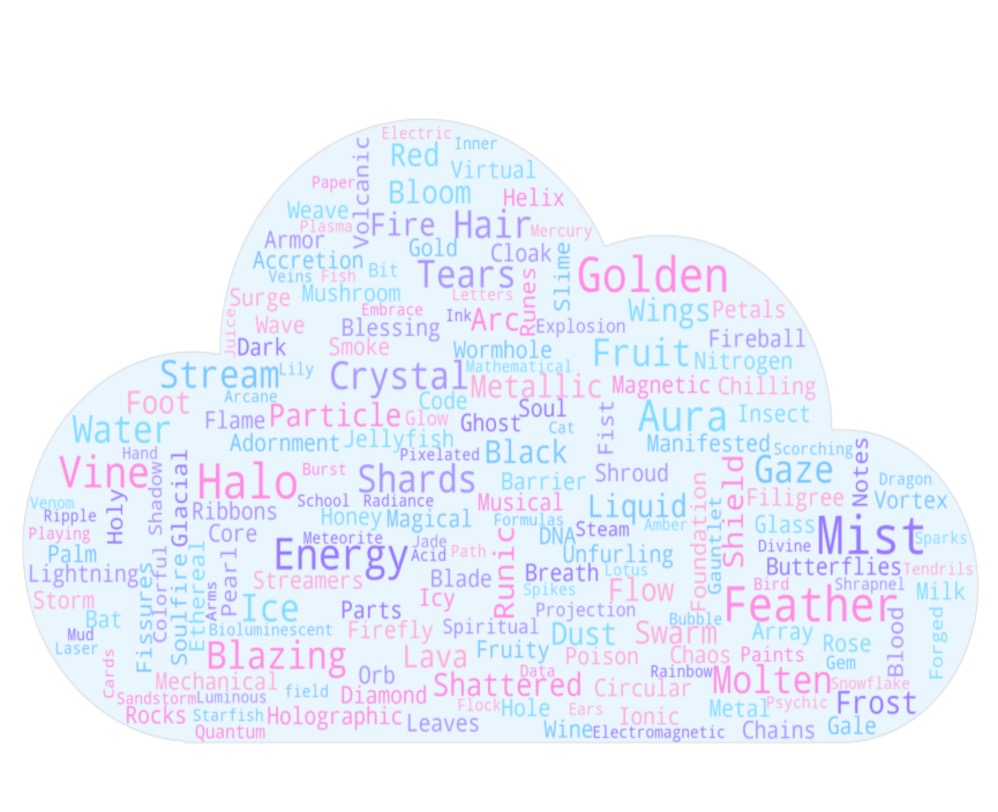}
    \caption{Word cloud of atmospheric VFX elements, showing the diversity of effect classes.}
    \label{fig:word_cloud}
\end{figure}

\section{Dataset details} \label{sec:sup.data}

Compared to conventional real-world video datasets, existing VFX datasets remain highly limited in both scale and diversity. To overcome this bottleneck, we develop a VFX synthesis pipeline that generates large-scale, high-quality paired data suitable for supervised VFX learning. The pipeline consists of five major steps.

\paragraph{Subject collection.}
We first construct a diverse subject dataset primarily containing human portraits and animal images. Human portraits are sourced from a mixture of internal datasets and publicly available collections such as PPR10K \cite{ppr10k}. Since these portraits are mostly female images, we additionally synthesize a subset of both male and animal images using FLUX \cite{flux}. Given the raw subject image pool, we perform three preprocessing steps: (1) Super-resolution: Images whose shorter side is below 704 pixels are upsampled via a high-fidelity super-resolution model. (2) Text erasure: We apply an OCR system to detect images with noticeable overlaid text and remove the text using Flux-Kontext \cite{kontext}. (3) Quality filtering: A VLM-based filter is used to discard multi-person images and samples where the primary subject occupies only a small portion of the frame, ensuring clean, centered subjects for downstream generation.

\paragraph{VFX taxonomy.}
To synthesize a broad spectrum of VFX types, we construct a systematic and structured VFX taxonomy. Overall, our effects fall into two major categories: atmospheric effects and transformation effects. 

Atmospheric effects refer to supernatural visual elements that originate from or surround the subject, enhancing mood, ambiance, or character attributes. To systematically describe such effects, we define orthogonal attribute sets that include: (1) VFX elements (\eg fire, ice, light), (2) geometric patterns (\eg, particles, waves, rings), and (3) attachment regions (\eg, face, ear, arms, full body). These attributes combine to form a rich space of atmospheric effects. \cref{fig:word_cloud} shows the word cloud of atmospheric effect group.

Transformation effects indicate substantial modifications to the subject’s appearance, including outfit transformation (\eg hairstyle, headwear, clothing, accessories \etc), identity transformation (\eg gender swap, aging, rejuvenation, or transformation into animals, robots, or mythical creatures), and physical transformation (\eg shattering, melting, explosion \etc).

\begin{figure*}[t]
  \centering
   \includegraphics[width=1.0\linewidth]{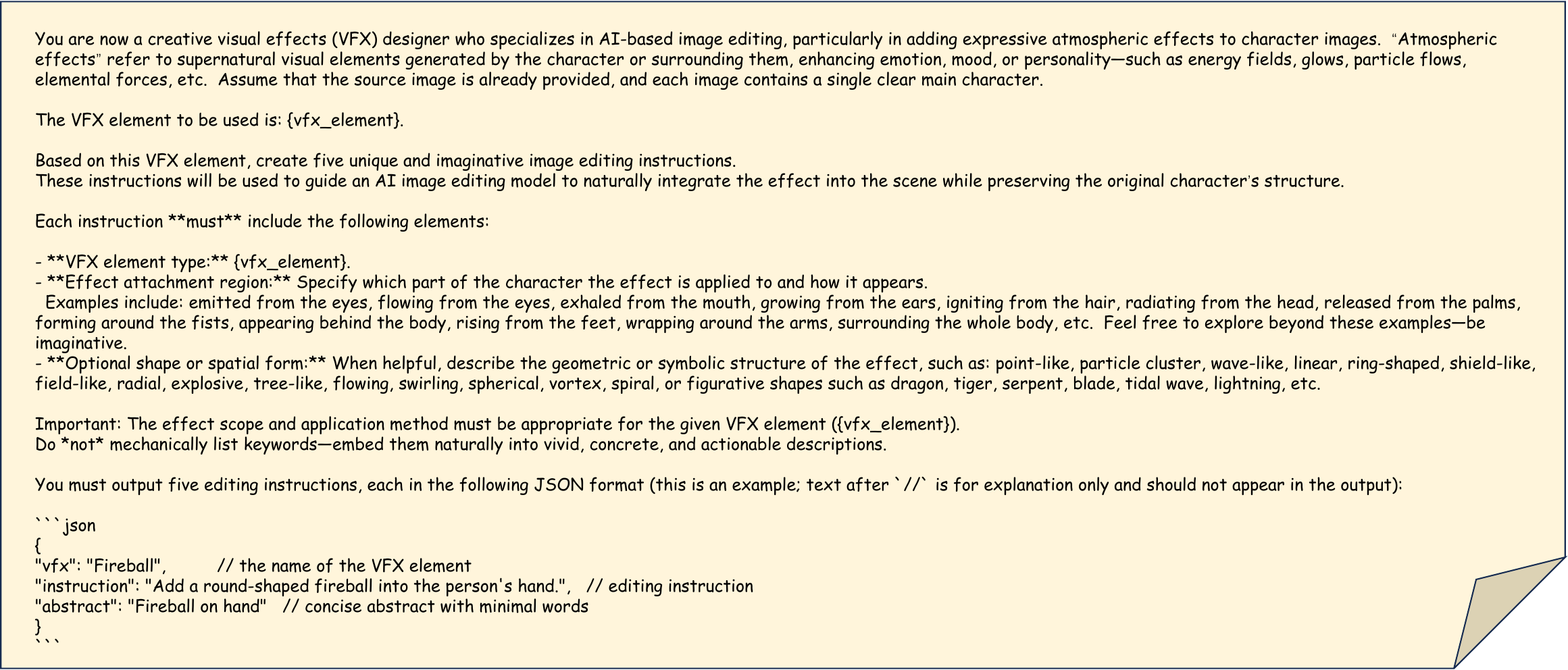}
    \caption{Prompt template for generating VFX editing instruction.}
    \label{fig:prompt_for_edit_instruction}
\end{figure*}

\paragraph{Instruction generation.}
For each atmospheric effect configuration, we prompt an LLM to combine the chosen effect element, geometric pattern, and attachment region into multiple diverse editing instructions that describe how to modify the source image into its VFX-enhanced counterpart. A prompt example is shown in \cref{fig:prompt_for_edit_instruction}.
For transformation effects, we similarly prompt the LLM to expand each basic transformation into a detailed editing instruction suitable for image editing models.

\begin{figure}[t]
  \centering
   \includegraphics[width=1.0\linewidth]{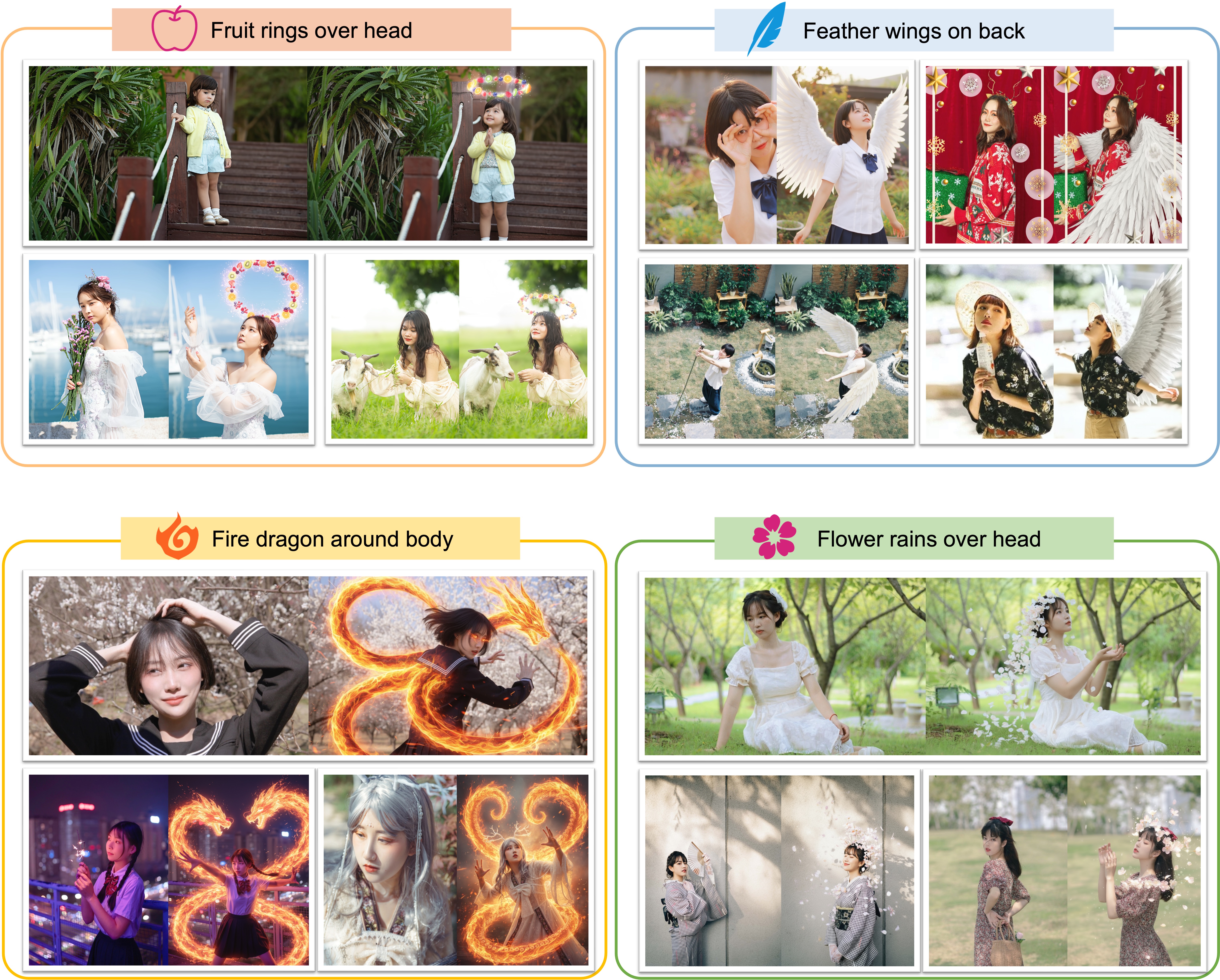}
    \caption{We show four groups of paired images under the same VFX. Each group contains three paired images with different subjects but the same effect.}
    \label{fig:data_sample}
\end{figure}

\paragraph{Subject editing.}
For each editing instruction, we randomly sample multiple subject images from the subject image bank. We then employ an image editing model to synthesize a target image that incorporates the desired effect. We experimented with several editing models, including Flux-Kontext \cite{kontext}, Qwen-Image-Edit \cite{qwenimage}, and NanoBanana \cite{banana}. These models can produce consistent VFX patterns across different subject images when given the same editing instruction, as shown in \cref{fig:data_sample}, which is crucial for generating coherent paired videos.
Beyond executing the core VFX instruction, we further prompt the model to adjust the subject’s pose, body motion, or facial expression to better match the semantics of the effect.
The edited image and its corresponding original images collectively form a paired VFX group, from which we can sample a reference-target sample pair for supervised in-context training.

\paragraph{Video generation.}
For each source–target pair, we horizontally concatenate the two images and feed the combined image along with the effect description into Gemini 2.5 Pro \cite{gemini}. We instruct the model to imagine the left image as the first frame and the right image as the last frame, and to describe the dynamic transition between them. 
Finally, we use the generated transition description together with the paired images as input to first–last-frame-to-video models, such as Jimeng3.0 \cite{jimeng} and Wan-FLF2V-14B \cite{wan} to synthesize complete VFX videos.

\paragraph{Scalable data generation.}
To reduce the cost of large-scale data synthesis with closed-source models, we adopt a scalable data expansion strategy. 
We first generate a moderate amount of high-quality data using the pipeline above and then fine-tune an instruction-based VFX generation model on an image-to-video backbone. 
The resulting model can produce consistent effects across different subjects given the same VFX instruction, enabling efficient large-scale data generation without repeatedly invoking expensive image editing and first–last-frame video models.






\section{Additional comparison} \label{sec:sup.compare}
\begin{figure*}[t]
  \centering
   \includegraphics[width=1.0\linewidth]{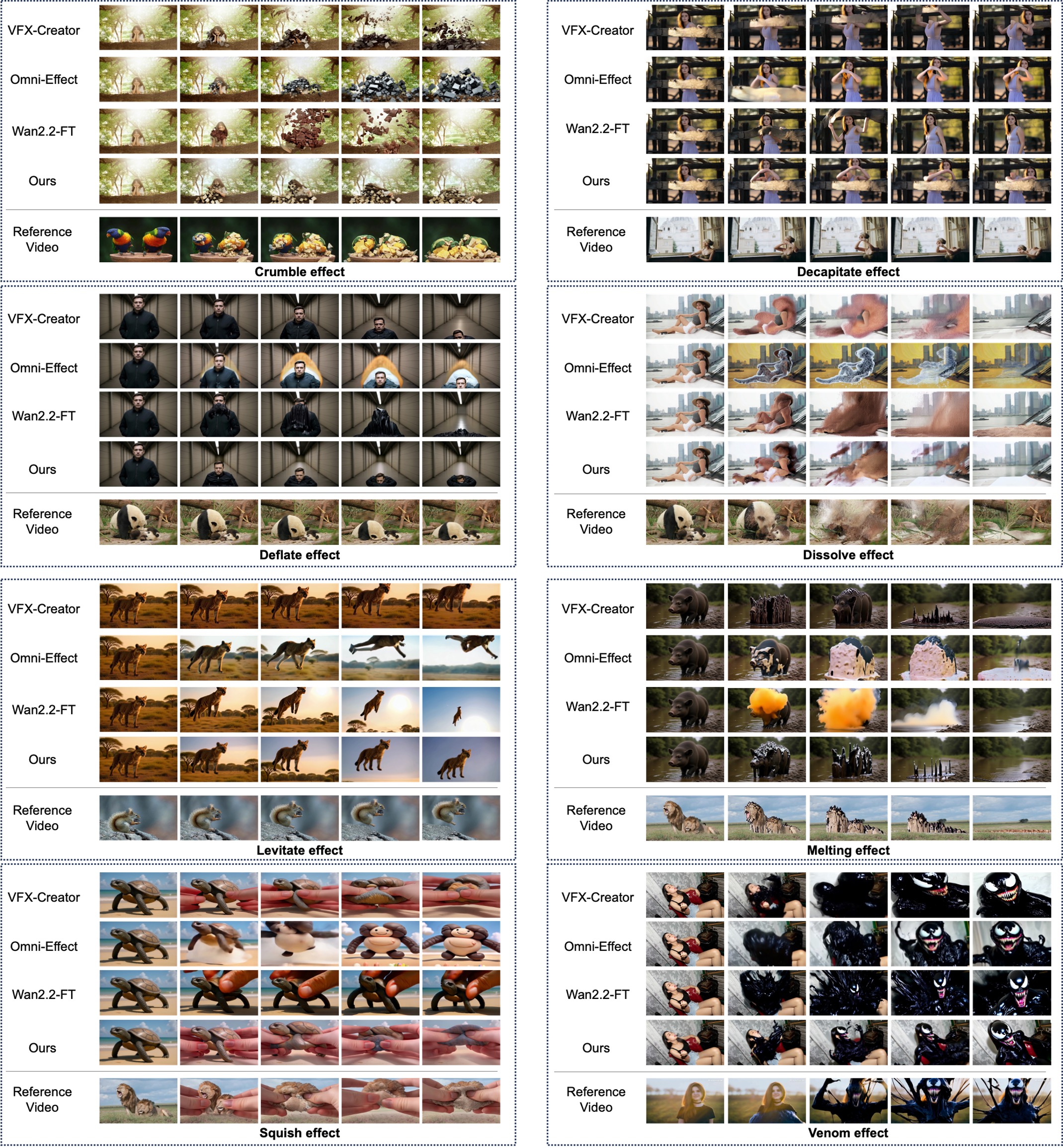}
    \caption{Additional visual comparison with related baselines on OpenVFX dataset.}
    \label{fig:openvfx_ad_compare}
\end{figure*}

We provide additional visual comparisons with related baselines on OpenVFX dataset in \cref{fig:openvfx_ad_compare}. Please refer to our project page for dynamic results.

\section{Additional visual results} \label{sec:sup.quali}
\begin{figure*}[t]
  \centering
   \includegraphics[width=1.0\linewidth]{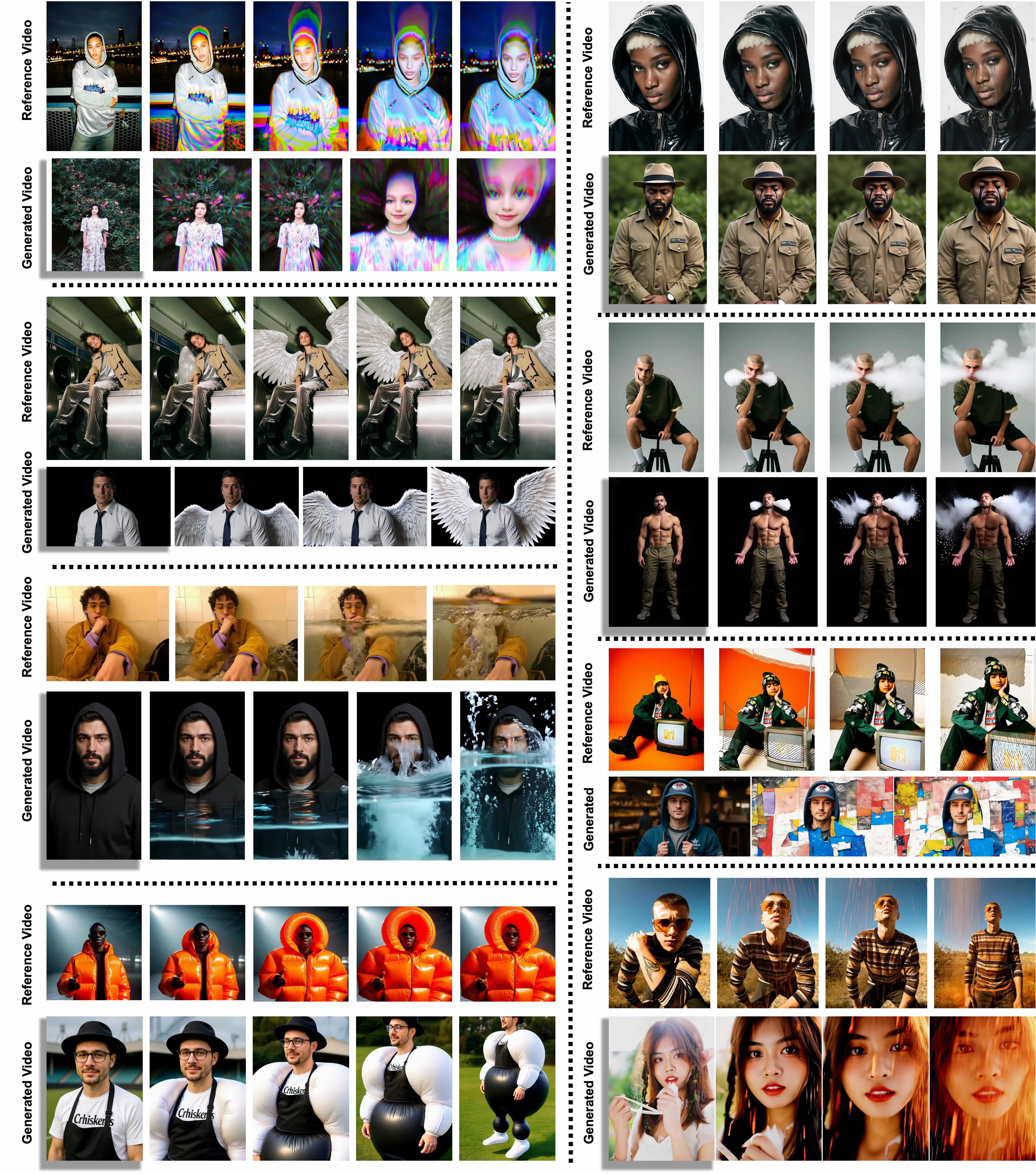}
    \caption{Additional visual results. In each grid, top row is the reference video, bottom row is our generated result.}
    \label{fig:q1}
\end{figure*}

\begin{figure*}[t]
  \centering
   \includegraphics[width=1.0\linewidth]{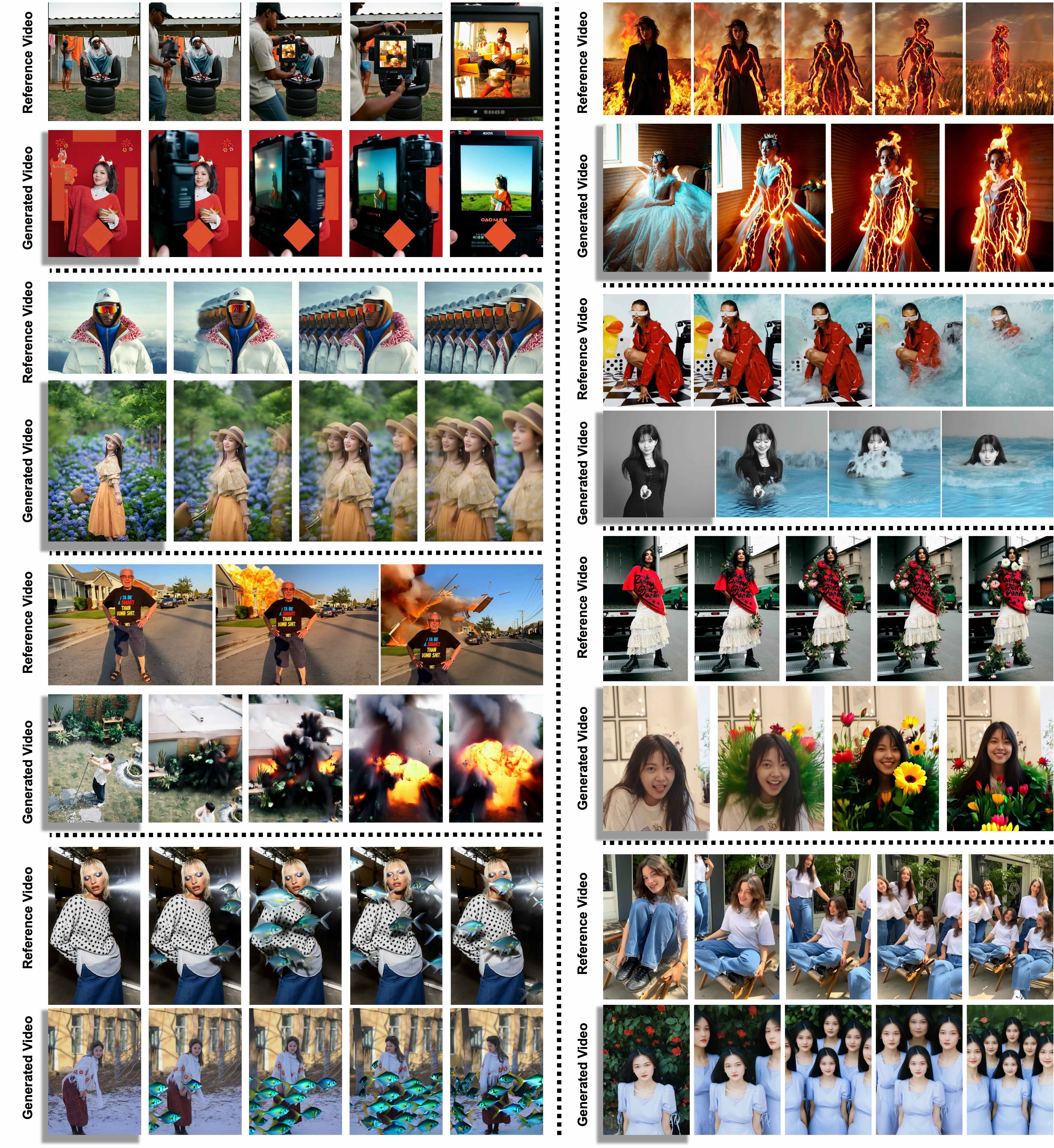}
    \caption{Additional visual results. In each grid, top row is the reference video, bottom row is our generated result.}
    \label{fig:q2}
\end{figure*}

\begin{figure*}[t]
  \centering
   \includegraphics[width=1.0\linewidth]{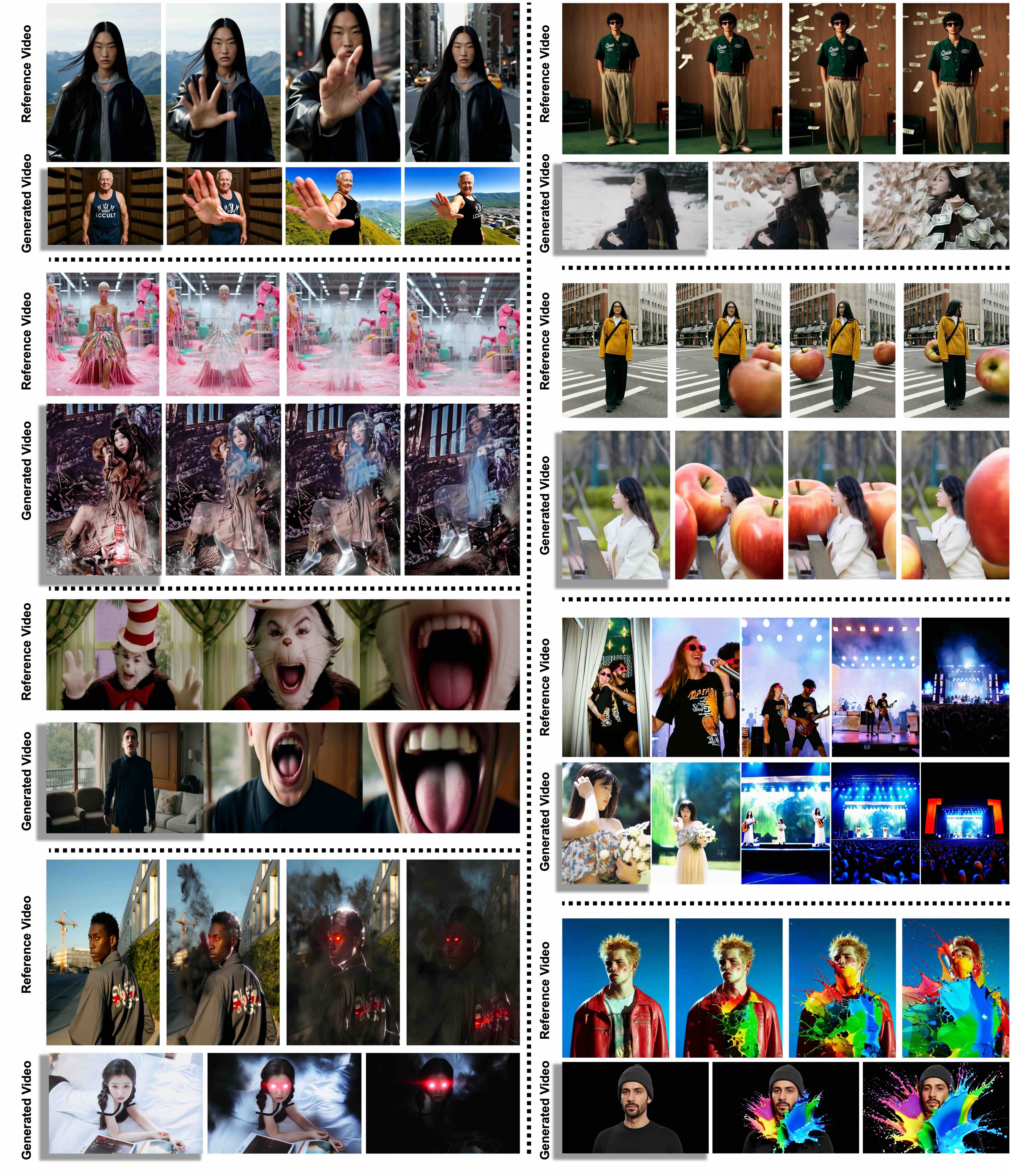}
    \caption{Additional visual results. In each grid, top row is the reference video, bottom row is our generated result.}
    \label{fig:q3}
\end{figure*}

\begin{figure*}[t]
  \centering
   \includegraphics[width=1.0\linewidth]{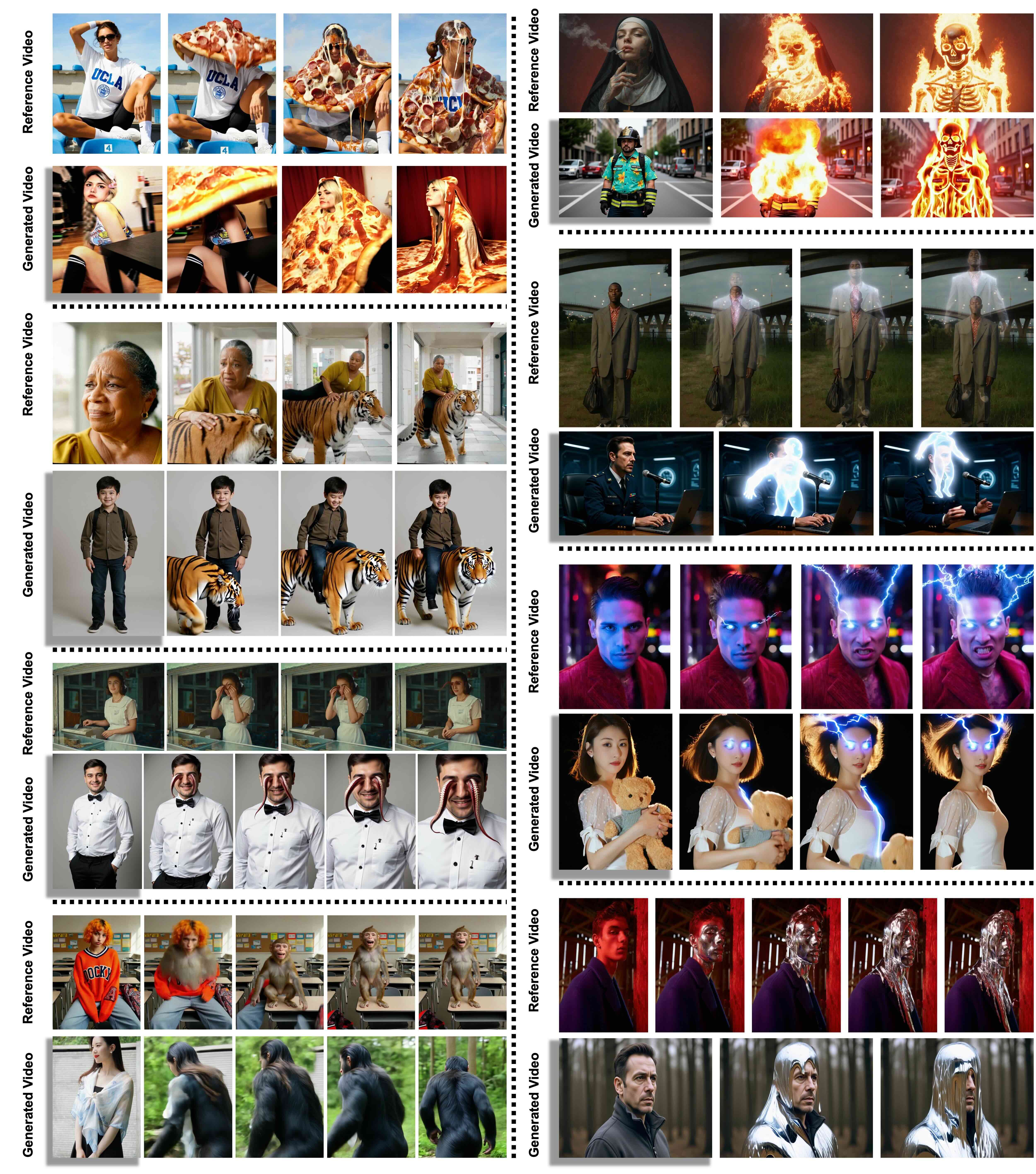}
    \caption{Additional visual results. In each grid, top row is the reference video, bottom row is our generated result.}
    \label{fig:q4}
\end{figure*}

\begin{figure*}[t]
  \centering  
   \includegraphics[width=1.0\linewidth]{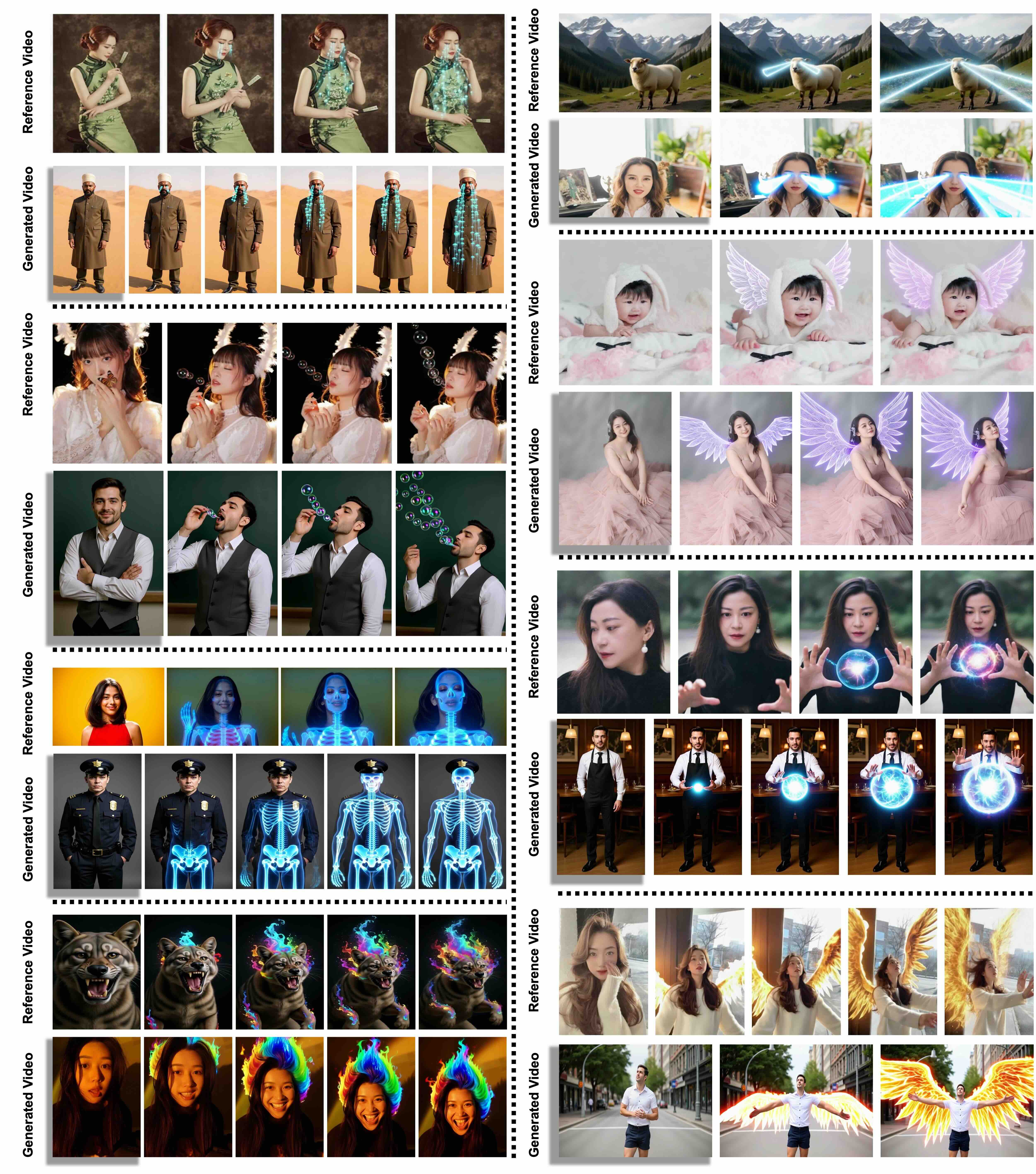}
    \caption{Additional visual results. In each grid, top row is the reference video, bottom row is our generated result.}
    \label{fig:q5}
\end{figure*}

We present additional VFX visual results other than OpenVFX dataset, including those from Higgsfield and EffectData, as shown in \cref{fig:q1}, \cref{fig:q2}, \cref{fig:q3}, \cref{fig:q4}, \cref{fig:q5}. Please refer to our project page for dynamic results. 

\section{Real-world VFX adaptation.}
\begin{figure*}[t]
  \centering
  \includegraphics[width=1.0\linewidth]{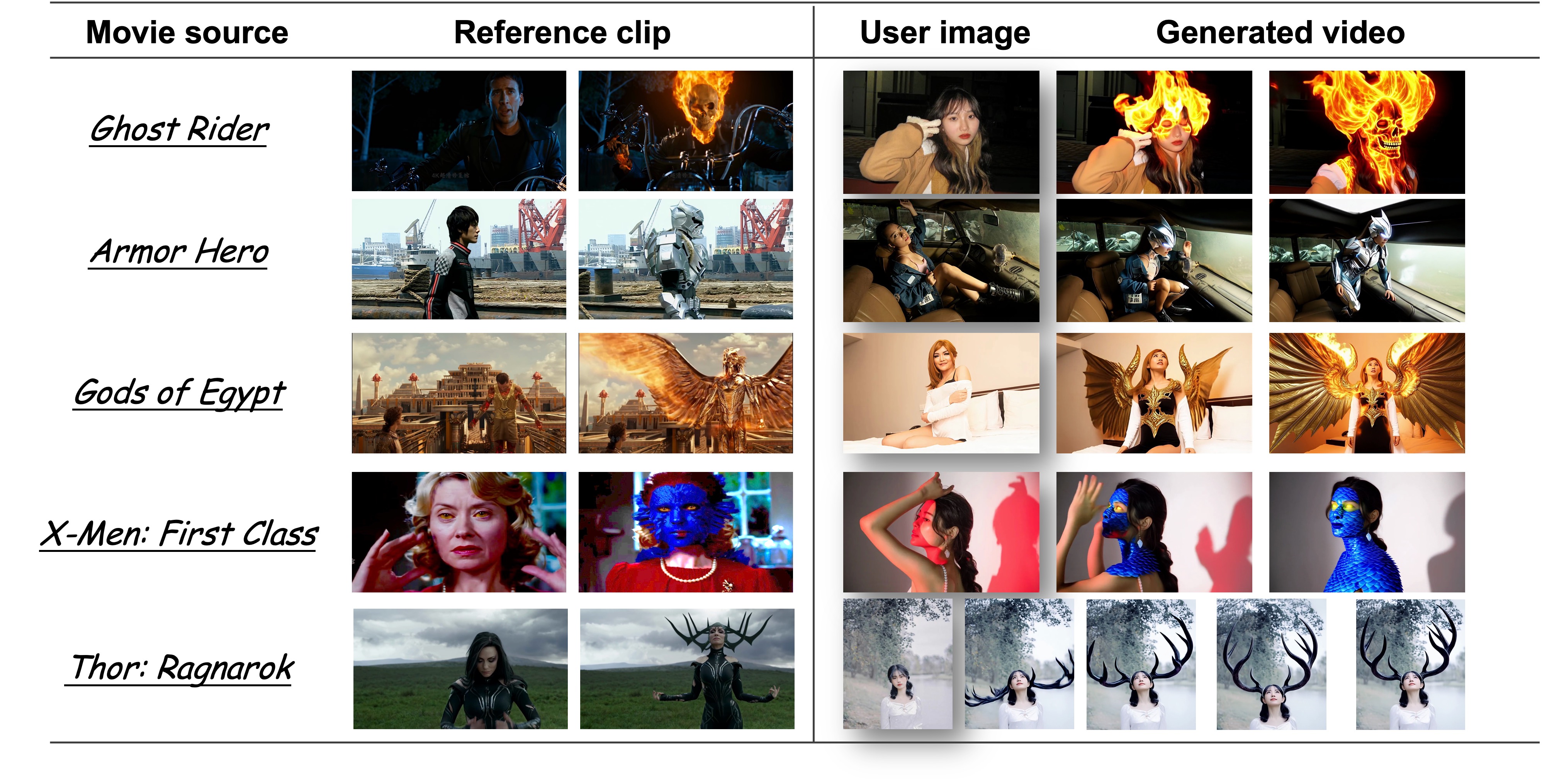}
   \caption{Real-world VFX transfer examples.}
   \label{fig:realvfx}
\end{figure*}

While our model was trained on synthetic VFX data, it still has the potential to generalize to real-world VFX transfer. We manually collected some real-world VFX movie clips and evaluated our model. As shown in Fig.~\ref{fig:realvfx}, despite being trained only on synthetic data, it achieves successful zero-shot generalization to real VFX. However, a synthetic-to-real gap remains in fine-grained details. We plan to bridge this in future work by incorporating real-world VFX data.

\section{Failure cases} \label{sec:sup.fail}
\begin{figure*}[t]
  \centering  
   \includegraphics[width=0.8\linewidth]{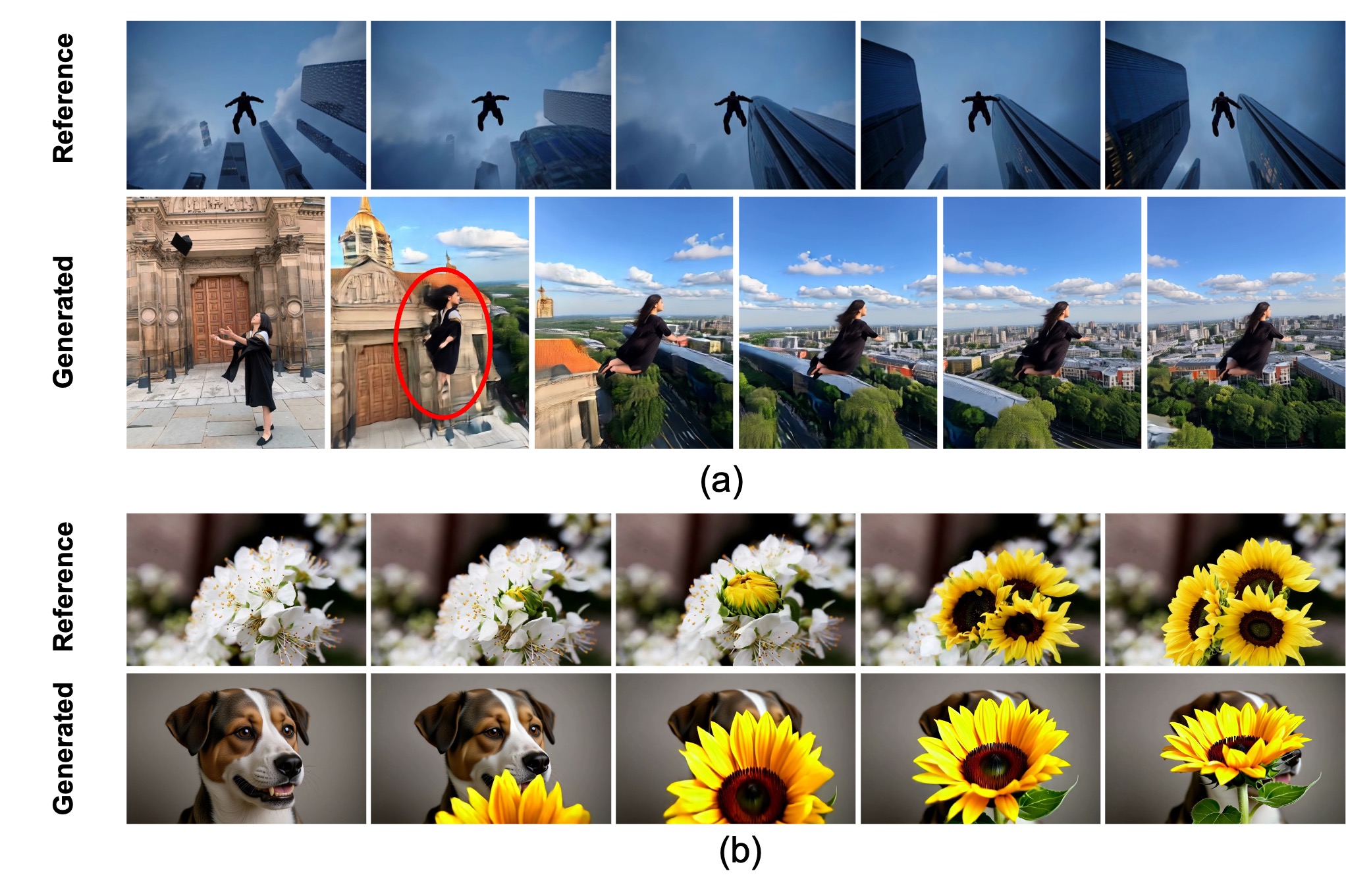}
    \caption{
    Failure cases of our method. 
    (a) Due to the limited capacity of the base model, subject fidelity may degrade under large and rapid motions (e.g., sudden upward take-off). 
    (b) When the user provides a first-frame image that is semantically incompatible with the reference effect, the transferred effect becomes incoherent and fails to reflect the intended VFX.}
    \label{fig:limit}
\end{figure*}

Despite the effectiveness of our proposed framework, there are still certain limitations and failure cases to be discussed.
First, due to the inherent capacity limitations of the base model, our approach may struggle with extremely complex effects involving rapid or large-magnitude motions. As shown in \cref{fig:limit}(a), when the reference effect is \textit{flying} and the subject undergoes a sudden take-off motion, the model occasionally fails to maintain subject fidelity (see red circle).
Second, the framework assumes that the provided first-frame image is semantically compatible with the reference effect. When this assumption is violated, the transferred effect may become incoherent or semantically meaningless. As illustrated in \cref{fig:limit}(b), the reference effect is \textit{sunflower blooming}, but the user supplies a dog image as the first frame. Under such mismatch, the generated video cannot faithfully manifest the intended VFX semantics.
Note that such issue can be avoided by proper user interactions. Our future work will employ more powerful base models and incorporate diverse real-world data sources to further enhance generalization and realism.

\end{document}